\documentclass{article}
\usepackage[T1]{fontenc}
\usepackage{iclr2026_conference,times}

\usepackage{amsmath,amsfonts,bm}

\def\eqref#1{equation~\ref{#1}}
\def\1{\bm{1}}

\DeclareMathAlphabet{\mathsfit}{\encodingdefault}{\sfdefault}{m}{sl}
\SetMathAlphabet{\mathsfit}{bold}{\encodingdefault}{\sfdefault}{bx}{n}

\usepackage{array,tabularx,booktabs}
\usepackage{xcolor,colortbl,pifont}
\usepackage{amsmath}
\usepackage{booktabs}
\usepackage{graphicx}
\usepackage{microtype}
\usepackage{multirow}
\usepackage{array}
\usepackage{xcolor}
\usepackage{enumitem}
\usepackage{hyperref}
\hypersetup{hidelinks}
\usepackage{url}
\usepackage{wrapfig}
\usepackage{capt-of}
\definecolor{ehrblue}{HTML}{4C78A8}
\definecolor{notegreen}{HTML}{59A14F}
\definecolor{ecgred}{HTML}{E15759}
\definecolor{cxrorange}{HTML}{F28E2B}
\definecolor{echopurple}{HTML}{B07AA1}

\newcommand{\benchmark}{\textbf{\texttt{ClinLens}}}
\newcommand{\strictpass}{\textsc{StrictPass}}
\newcommand{\finalacc}{\textsc{FinalAcc}}
\newcommand{\execsuccess}{\textsc{ExecSuccess}}

\newcommand{\eat}[1]{}

\title{ClinLens: Towards Long-Horizon LLM Agents for Longitudinal Multimodal Clinical Data Science}

\author{\textbf{Yuan Zhu}$^{1,\ast}$ \quad
\textbf{Ethan B. Liu}$^{1,\ast,\dagger}$ \quad
\textbf{Frank Nie}$^{1,\ast}$ \quad
\textbf{Wei Fan}$^{2}$ \quad
\textbf{Haibo Pu}$^{3}$ \quad
\textbf{Jindong Han}$^{1,\dagger}$ \\[4pt]
{\normalfont\small $^{1}$ Shandong University, China} \\
{\normalfont\small $^{2}$ The University of Auckland, New Zealand} \\
{\normalfont\small $^{3}$ Karlsruhe Institute of Technology, Germany} \\
{\normalfont\small \texttt{jindong.han@sdu.edu.cn}}
}

\iclrfinalcopy % 显示真实作者，同时关闭审稿行号
\begin{document}

% Keep the template's fonts, margins and line spacing; do not stretch
% heading/paragraph gaps merely to align the bottoms of float-heavy pages.
\raggedbottom

\maketitle
\fancyhead{}
% !TEX root = ../main.tex

\begin{abstract}
Automating clinical data science requires agents to translate patient-centric objectives into executable workflows grounded in longitudinal multimodal evidence. Existing benchmarks provide limited coverage of two essential capabilities: grounding analyses in the correct patient episode and harmonizing multimodal observations. We introduce \benchmark{}, a patient-centric benchmark for evaluating large language model (LLM) agents on long-horizon clinical data science workflows. \benchmark{} comprises 126 executable tasks spanning five data modalities, four analytical scopes, and five analytical capabilities. The tasks assess agents' ability to maintain a coherent patient-specific analytical context across interdependent stages of evidence grounding, multimodal integration, and analytical pipeline generation and execution. We evaluate general-purpose, coding, and biomedical agents, with task runs averaging 36.93 steps and 8.06 minutes. The strongest evaluated agent achieves an execution success rate of 86.4\% and a final-answer accuracy of 54.0\%, but a strict pass rate of only 29.4\%. Failure analysis identifies patient-episode grounding as the most frequently recorded failure category under the evaluation procedure. By jointly evaluating analytical outcomes and the validity of the supporting clinical context, \benchmark{} provides a testbed for advancing reliable agents for longitudinal clinical data science.
\end{abstract}

% !TEX root = ../main.tex

\section{Introduction}
% Longitudinal clinical data include multimodal records (e.g., structured tables, physiological time series, medical images) collected across hospital admissions and care settings, which is valuable for characterizing disease progression and treatment response of individual patients~\cite{johnson2023mimic}. Conducting computational analysis over such data often requires a series of analytical and coding steps, such as identifying relevant patient records and clinical episodes, harmonizing multimodal data sources, performing statistical analysis, and building predictive models. Recent advances in large language models (LLMs), particularly in complex reasoning~\cite{guo2025deepseek} and code generation~\cite{novikov2025alphaevolve}, create new opportunities to automate clinical data science workflows~\cite{jing2025dsbench,zhang2026deepanalyze}. However, reliably translating patient-specific analytical tasks into executable computational solutions grounded in longitudinal clinical evidence remains a significant challenge.
Clinical data science often requires computational analysis for individual patients to characterize disease progression and treatment response. Such analysis draws on longitudinal clinical data collected across admissions and care settings, e.g., structured records, physiological time series, and medical images~\citep{johnson2023mimic}. To conduct analysis, researchers typically construct a long-horizon data science workflow, including identifying relevant clinical episodes, integrating multimodal data, and implementing statistical or predictive methods. Recent advances in large language models (LLM), particularly in reasoning~\citep{guo2025deepseek} and code generation~\citep{novikov2025alphaevolve}, offer new opportunities for agents to automate these workflows~\citep{jing2025dsbench,zhang2026deepanalyze}. However, reliably translating patient-specific analytical objectives into executable workflows grounded in longitudinal clinical evidence remains a central challenge.

% Designing LLM agents for longitudinal clinical analysis poses unique challenges beyond general-purpose medical reasoning and code generation. Existing benchmarks evaluate important aspects of medical reasoning~\cite{wang2026liveclin}, biomedical data analysis~\cite{xu2026medagentgym,wang2026making}, and executable data science workflows~\cite{liu2026moscibench}, but generally organize tasks around individual modalities or datasets rather than longitudinal, multimodal patient trajectories. This leaves two critical evaluation gaps: (1) \emph{patient-centric longitudinal grounding}, which requires linking clinical evidence to the correct patient and clinical episode across admissions, ICU stays, and repeated studies; and (2) \emph{asynchronous multimodal integration}, which demands aligning heterogeneous clinical modalities acquired at different times and sampling frequencies within task-specific temporal windows.

Evaluating this capability requires benchmarks centered on longitudinal, multimodal patient trajectories. Existing benchmarks mainly focus on evaluating agents on automating dataset-centric analytical workflows~\citep{xu2026medagentgym,wang2026making,xu2026longmedbench}, but provide limited coverage of patient-centric evidence integration across clinical episodes and modalities. Specifically, two key capabilities remain insufficiently evaluated: (1) \emph{patient-episode grounding}, which links evidence to the correct patient and clinical episode across admissions, ICU stays, and repeated studies; and (2) \emph{asynchronous multimodal integration}, which aligns multimodal observations collected at different times and sampling rates within task-specific temporal windows.

To address this gap, we introduce \benchmark, a patient-centric benchmark for evaluating LLM agents on longitudinal clinical data science. Our contributions are threefold:
\begin{itemize}[leftmargin=*, labelindent=0pt]
\item \textbf{Problem framing}. We formulate longitudinal clinical analysis as an end-to-end workflow connecting patient-episode grounding and asynchronous multimodal integration with analytical pipeline generation, execution, and verification. This formulation makes constructing a valid patient-centric analytical context an explicit task requirement.
\item \textbf{Benchmark}. We instantiate this formulation in 126 executable tasks spanning five data modalities, i.e., structured electronic health records (EHR), clinical notes, electrocardiogram (ECG), chest X-ray (CXR), and echocardiogram (ECHO). A 4 $\times$ 5 task taxonomy crosses four patient-episode levels (patient, admission, ICU stay, and study) with five clinical analysis capabilities, i.e., course profiling, correlation analysis, treatment effect estimation, risk prediction, and phenotype discovery. In practice, each task runs averaging 36.93 steps and 8.06 minutes.
\item \textbf{Empirical findings}. We evaluate a broad range of general-purpose, coding, and biomedical agents, revealing a substantial gap between execution, answer correctness, and workflow validity. The strongest evaluated agent achieves 86.4\% execution success and 54.0\% final-answer accuracy, but only a 29.4\% strict pass rate. Failure analysis further identifies patient-episode grounding as the most frequently observed failure under the evaluation procedure.
\end{itemize}

\begin{figure}[t]
  \centering
  \includegraphics[width=0.8\linewidth]{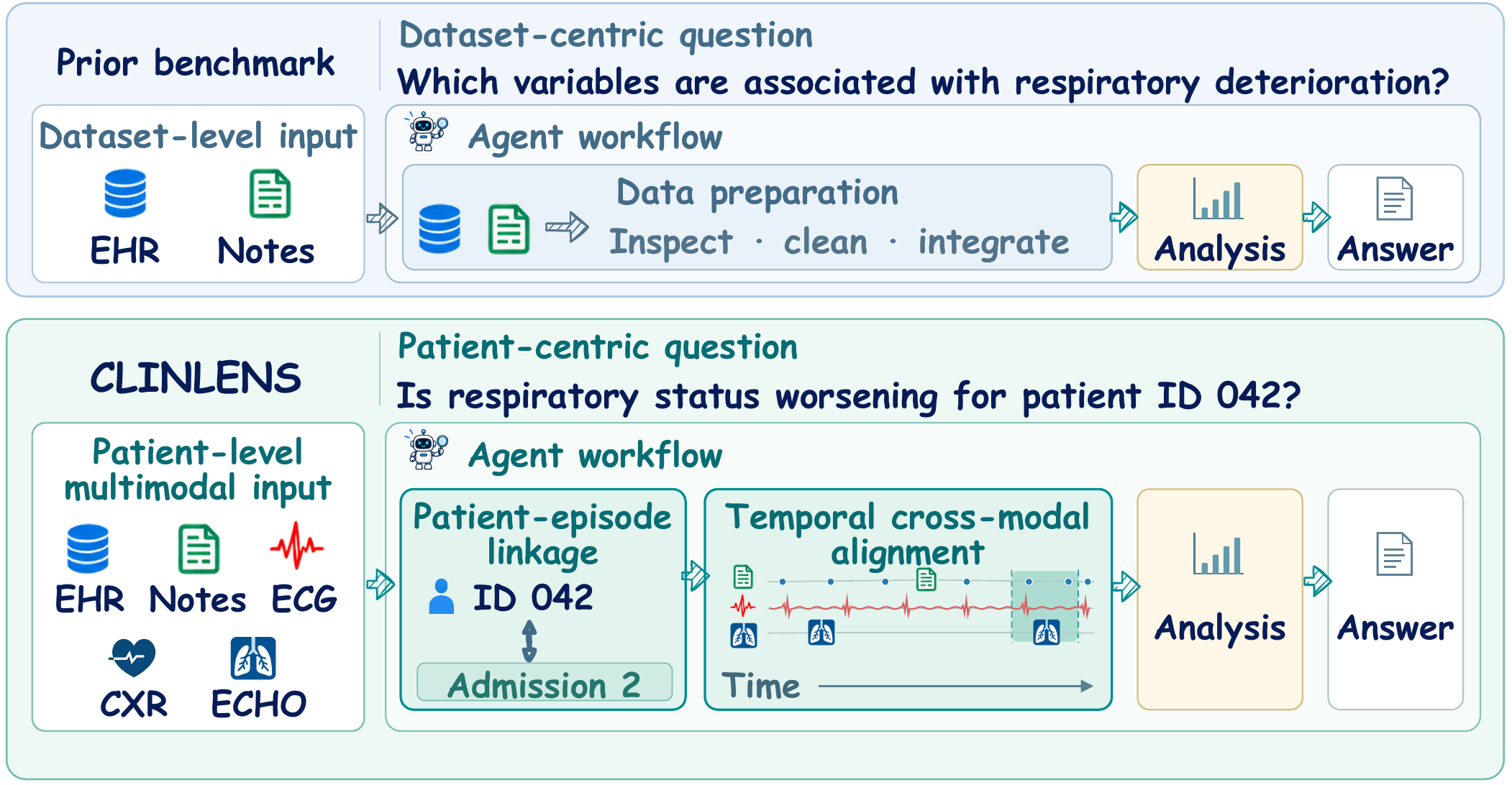}
  \caption{\textbf{Prior benchmarks vs. \benchmark.}
  Left: Existing benchmarks mainly focus on dataset-centric tasks with limited modality coverage. Right: \benchmark{} evaluates patient-centric clinical analysis across five modalities, requiring agents to link evidence to the correct patient and episode, align asynchronous multimodal observations before generating and executing analytical workflows.}
  \label{fig:comparison}
\end{figure}

% \begin{samepage}
% \noindent Our contributions are threefold:
% \begin{itemize}[leftmargin=1.3em,itemsep=2pt,topsep=3pt]
% \item \textbf{Problem framing.} We frame patient-centered analysis as an end-to-end workflow linking cohort construction, identity resolution, temporal alignment, leakage control, and result verification.
% \item \textbf{Benchmark.} We provide 20 workflow templates across five MIMIC modalities and four patient-time levels, with executable reference workflows and deterministic artifact-level grading.
% \item \textbf{Empirical findings.} Comparisons of general-purpose, coding, and biomedical agents expose gaps between execution, answer correctness, and workflow validity, with episode reconstruction a principal observed bottleneck across the non-prediction diagnostic cohort.
% \end{itemize}
% \end{samepage}
% We release the task-construction and evaluation code and plan to distribute derived packages through credentialed PhysioNet access under the source-data agreements and release restrictions.

% !TEX root = ../main.tex
% Requires graphicx and capt-of (or caption), already present in the supplied project.

\section{Benchmark Construction}
\label{sec:benchmark-construction}
\begin{figure}[t]
  \centering
  \includegraphics[width=\linewidth]{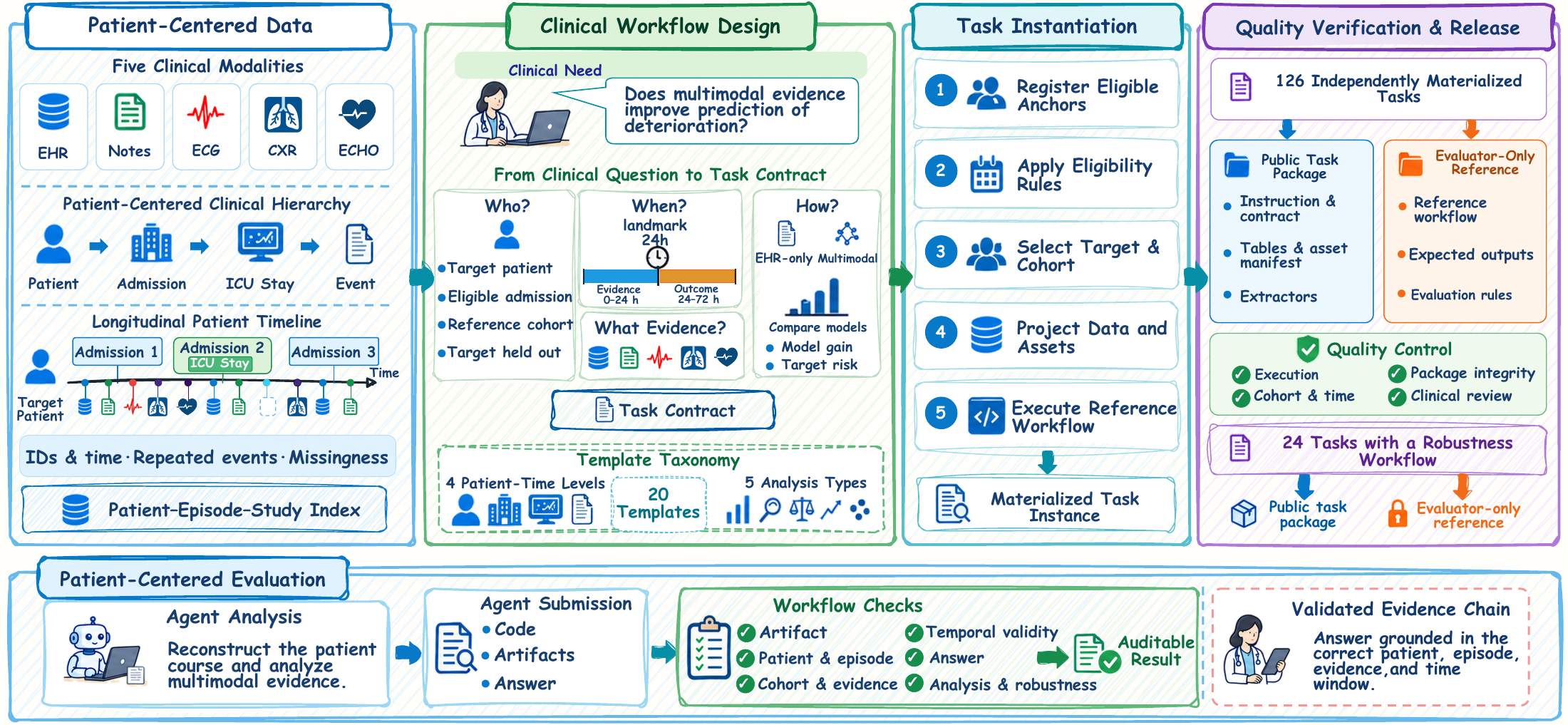}
  \caption{\textbf{Construction and evaluation pipeline of \benchmark{}.} Five clinical data sources are organized into patient-centric packages. Workflow templates and executable references define the task instances, while submitted artifacts and final answers support deterministic evaluation.}
  \label{fig:pipeline}
\end{figure}

\benchmark{} evaluates whether agents can reconstruct patient-episode relationships, align asynchronous multimodal evidence, and produce an auditable clinical analysis. Its construction connects clinical questions, executable reference workflows, and artifact-level evaluation (Figure~\ref{fig:pipeline}).

\subsection{Task Definition and Evaluation}
\label{sec:task-contract}

\paragraph{Task instructions.}
Each task asks an agent to answer a clinical data-analysis question for a named target patient within a specified longitudinal setting. The instruction defines the analysis population, temporal constraints, relevant modalities, required computation, and output contract. Depending on the question, the analysis may concern the target patient's course or involve estimation and modeling using a reference cohort. The agent must implement the analysis and submit a final answer together with required artifacts, such as cohort tables, temporal alignments, statistics, or model outputs. These artifacts allow the evaluator to inspect how the answer was obtained.

\paragraph{Longitudinal multimodal patient packages.}
Each task provides a bounded package containing the target patient and a reference cohort, with structured EHR records, notes, ECGs, CXRs, and ECHOs as relevant to the task. The package preserves native patient and episode identifiers, timestamps, repeated observations, and modality missingness rather than providing a pre-aligned feature table. Agents must link records to the correct patient and clinical episode, align asynchronously recorded modalities within the specified windows, and perform the necessary preprocessing and analysis. Thus, patient-centric grounding establishes whose evidence is used and which episode it belongs to, while temporal alignment establishes how observations from different modalities can be combined.

\paragraph{Evaluation.}
Each task is paired with an evaluator-private executable reference that produces the expected artifacts and answer under fixed analytical and temporal specifications. Evaluation first checks artifact completeness and parseability, then validates patient, cohort, temporal, and analytical semantics, and finally compares the submitted answer with the reference under task-specific matching rules. Final-answer correctness and full workflow validity are assessed separately: a matching answer alone does not establish that the underlying analysis satisfies the task contract. Operational details and metric definitions appear in Appendices~\ref{app:construction-details} and~\ref{app:evaluation-protocol}.

\subsection{Construction Pipeline and Coverage}
\label{sec:construction-coverage}
\paragraph{Patient-centric data curation.}
We integrate MIMIC-IV structured records~\citep{johnson2024mimiciv,johnson2023mimicivpaper}, MIMIC-IV-Note~\citep{johnson2023mimicnote}, MIMIC-IV-ECG~\citep{gow2023mimicecg}, MIMIC-CXR-JPG~\citep{johnson2024mimiccxrjpg}, and MIMIC-IV-ECHO~\citep{gow2026mimicecho}. Shared patient identifiers assemble bounded packages retaining native admission, ICU-stay, study, and event identifiers. Source timestamps, repeated examinations, one-to-many relations, and modality missingness remain visible, as do modality-specific representations: structured tables, narrative notes, physiological waveforms, radiographs, and echocardiographic studies.

We standardize schemas and file paths and register accessible raw assets, but do not precompute task-specific episode assignments, cross-modal joins, temporal filtering, repeated-measure aggregation, or analytical features. In particular, linking a study without an encounter identifier to the appropriate episode remains an agent responsibility, as does handling missing observations. This boundary standardizes data access while retaining the patient-grounding and temporal-alignment work being evaluated.

\begin{figure}[t]
  \centering
  \includegraphics[width=\linewidth]{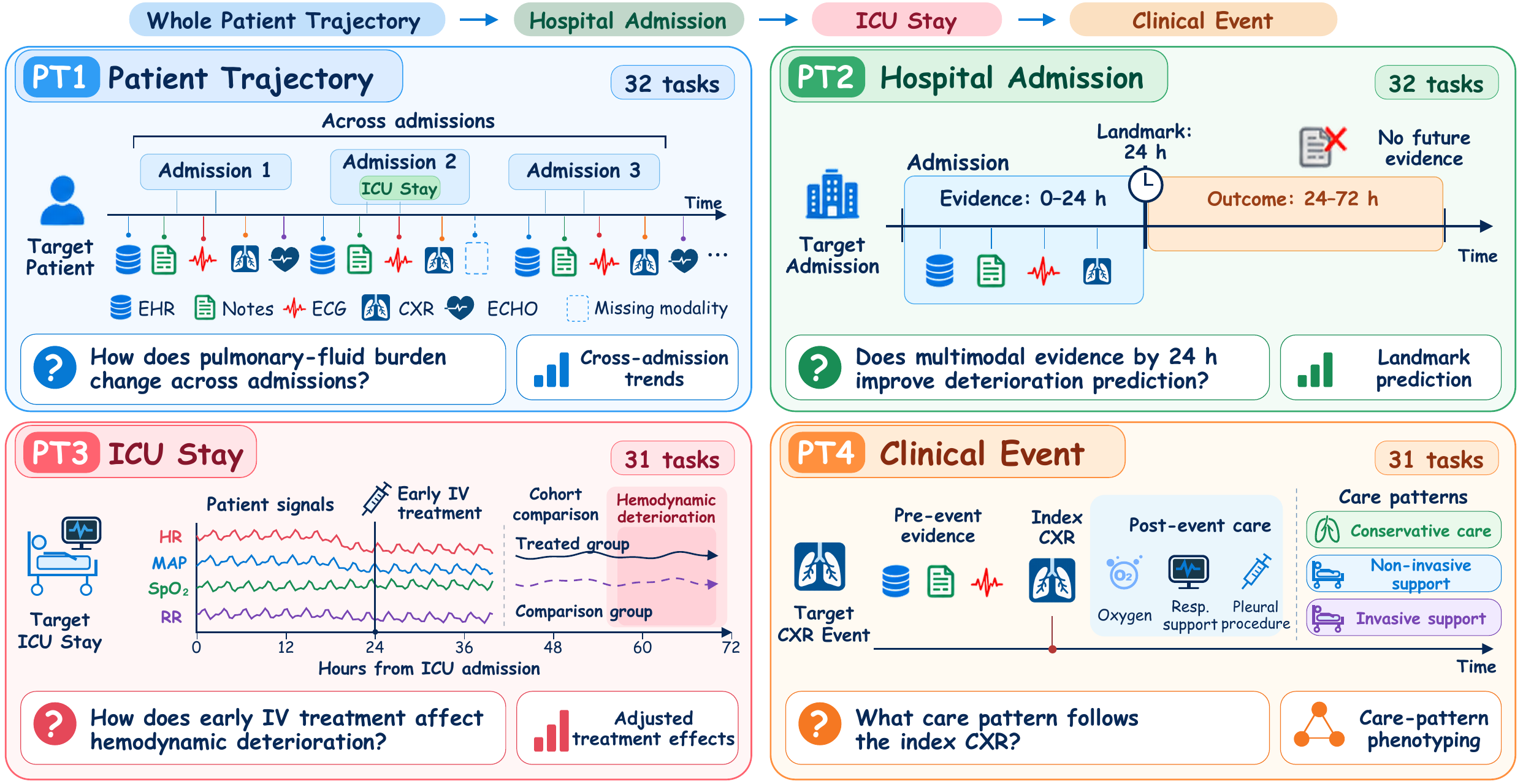}
  \caption{\textbf{Representative tasks across four patient-time scopes.} PT1 follows changes across admissions; PT2 uses admission-landmark evidence for prediction; PT3 estimates effects within an ICU stay; and PT4 analyzes patterns around a clinical event or study. Examples illustrate individual tasks, not all 20 templates. Timelines and physiological traces are schematic.}
  \label{fig:taxonomy}
\end{figure}

\paragraph{Clinical design and task instantiation.}
Construction begins with a clinical question. A workflow template fixes the eligible population, clinical anchor, evidence and outcome windows, relevant modalities, analytical objective, and expected outputs. For example, T09 compares EHR-only and multimodal models for subsequent deterioration using evidence available within the first 24 hours of an admission. Answering it requires both correct admission linkage and temporally eligible evidence before modeling. Fixed selection and sampling rules bind each template to eligible targets and independently materialized packages, yielding 126 tasks across the 20 templates. For each instance, an executable reference implements the question under prespecified preprocessing, analysis, and fallback rules. Its execution checks feasibility and generates the expected artifacts and answer; the instruction, artifact requirements, and evaluation rules are derived from this validated protocol.

\paragraph{Quality control and reference isolation.}
Template-level clinical and statistical review examines cohort definitions, temporal directionality, modality use, variable definitions, and analytical procedures. Each materialized instance is rerun in a clean environment to check package boundaries, raw-asset accessibility, identifier and timestamp consistency, artifact schemas, row-level and temporal semantics, and reference-artifact--answer agreement. Only instances whose packages, reference workflows, and expected submissions pass the complete evaluator are retained. Public task packages are separated from evaluator-private executable references and expected artifacts and answers. This separation permits reproducible grading without exposing the reference solution to agents. Appendix~\ref{app:construction-details} provides the operational details.

% Keep taxonomy text and its figure together; total width is 0.97\linewidth.
% The left block contains the coverage description, not unrelated QC text.

\subsection{Benchmark Statistics}
\label{sec:benchmark-statistics}

% Start the statistics block AFTER its full-width subsection heading.
% Two aligned minipages avoid a wrapfigure crossing the heading or leaving
% the final lines of one paragraph to switch from narrow to full width.
\noindent
\begin{minipage}[t]{0.53\linewidth}
\vspace{0pt}
Code-grounded annotation of the reference workflows identifies a mean of 36.93 analytical operations per task (range: 17--63). These are logical analysis steps, not model interaction turns: they cover patient identification, episode construction, multimodal temporal alignment, statistical analysis, and validation.

Figure~\ref{fig:solution-time} reports task-level mean solving times, obtained by averaging available model--agent runtimes within each task. The overall mean is 8.06 minutes; capability means range from 5.34 minutes for course profiling to 10.02 minutes for association analysis. Runtimes include unsuccessful attempts and depend on the evaluated systems. They describe observed execution cost, whereas operation counts characterize the structure of the reference workflows.
\end{minipage}\hfill
\begin{minipage}[t]{0.44\linewidth}
\vspace{0pt}
\centering
\includegraphics[width=\linewidth]{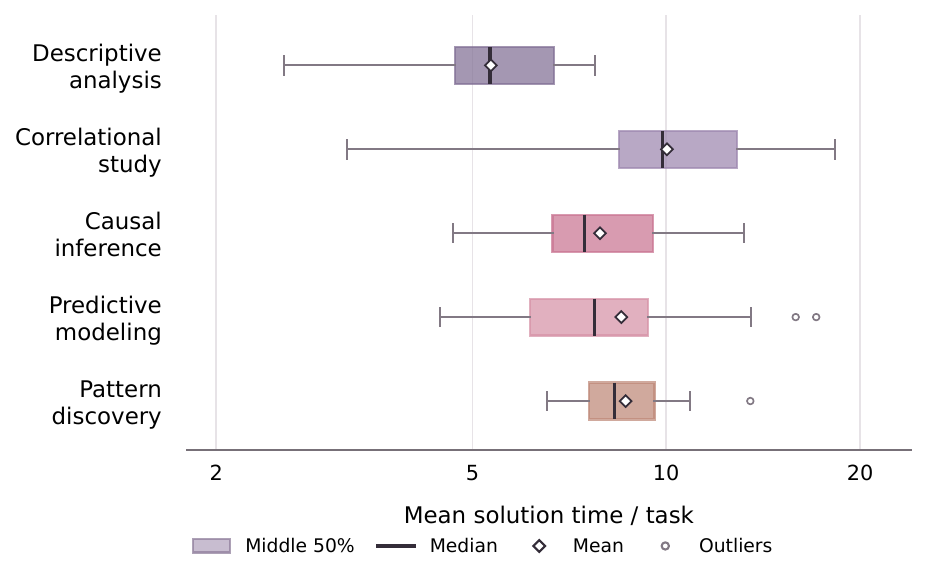}
\captionof{figure}{\textbf{Task-level mean solving time by clinical capability.}}
\label{fig:solution-time}
\end{minipage}
\par

\par\noindent
\begin{minipage}[t]{0.53\linewidth}
\vspace{0pt}
\textbf{Workflow taxonomy.}
Four patient-time scopes define the analysis unit: whole-patient trajectories, hospital admissions, ICU stays, and clinical event- or study-centered windows, containing 32, 32, 31, and 31 tasks, respectively. Figure~\ref{fig:taxonomy} illustrates a representative task at each scope.
Five clinical capabilities define the analytical objective: \emph{clinical course profiling} (descriptive analysis), \emph{clinical association analysis} (correlational analysis), \emph{treatment effect estimation} (causal inference), \emph{clinical risk prediction} (predictive modeling), and \emph{clinical phenotype discovery} (pattern discovery). Crossing the two axes produces 20 workflow templates. The suite contains 26 profiling tasks and 25 tasks in each other capability (Figure~\ref{fig:category-distribution}). Each task uses a clinically relevant subset of the five modalities; the complete template index and modality coverage remain in Appendix~\ref{app:templates} and Table~\ref{tab:coverage}.
\end{minipage}\hfill
\begin{minipage}[t]{0.44\linewidth}
\vspace{0pt}
\centering
\includegraphics[width=\linewidth]{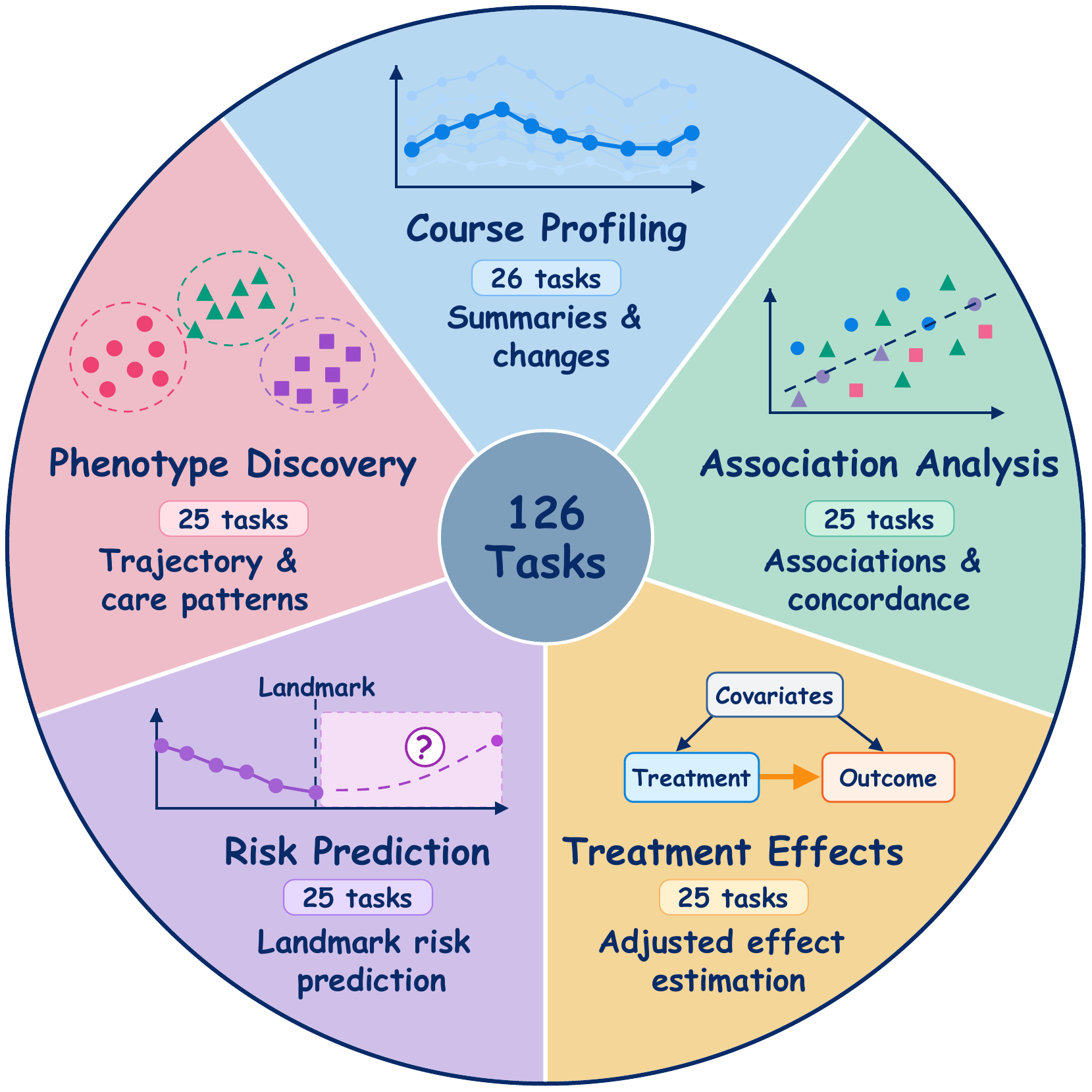}
\captionof{figure}{\textbf{Clinical capability taxonomy and task counts.}}
\label{fig:category-distribution}
\end{minipage}
\par\medskip

% \input{sections/evaluation_protocol}
% \input{sections/results}
% !TEX root = ../main.tex

\section{Experiments}
\label{sec:experiments}
We organize the evaluation around four research questions:
\begin{itemize}[leftmargin=*,itemsep=1pt,topsep=3pt]
\item \textbf{RQ1 (Capability):} How does performance vary across capabilities and patient-time levels?
\item \textbf{RQ2 (Failure localization):} Where do workflows first violate the clinical analysis contract?
\item \textbf{RQ3 (Answer validity):} How often is a correct answer supported by a fully valid workflow?
\item \textbf{RQ4 (Efficiency):} How do attempts and tokens relate to answer and workflow validity?
\end{itemize}

\subsection{Experimental Setup}
\label{sec:experimental-setup}
We evaluate a fixed suite of 126 clinical data-analysis tasks. Its selection emphasizes historically difficult workflows, so scores characterize this stress suite rather than an IID sample of clinical analyses. Standard configurations use zero-shot prompting, temperature 0, and a one-hour execution limit. Access is restricted to assigned packages and listed raw assets; private references and previous submissions are withheld. Frameworks, iteration limits, and metric thresholds are detailed in Appendix~\ref{app:evaluation-protocol}.

The system comparison includes standardized scaffolds, coding harnesses, and biomedical-agent adapters (Table~\ref{tab:main-results}). DeepSeek-V4-Pro and GLM-5.2 retain evaluator-compatible historical results; Qwen-3.8-Max counts unevaluated tasks as failures. The diagnostic analyses use ten complete data-grounded configurations: GPT-5.6-Luna and Gemini-3.7-Flash, each paired with ReAct, DataVoyager, Reflexion, SelfDebug, and RAG-ReAct. NoDataGuess, coding harnesses, biomedical-agent adapters, and configurations with historical results or incomplete coverage are excluded. These analyses therefore characterize the selected standard data-grounded configurations rather than the full system comparison.

We report \finalacc{} for final-result agreement, \strictpass{} for agreement plus all required artifact, schema, row-semantic, and validation gates, and \execsuccess{} for completed execution with an accepted submission. The 24 robustness tasks require both base and additional analyses. \emph{Artifact-ready} checks completeness and schema from evaluator reports and differs from \execsuccess{}. System and efficiency tables macro-average patient-time levels; diagnostics pool 1,260 records for RQ1/RQ3 and 1,010 non-prediction records for RQ2 (Table~\ref{tab:analysis-populations}). An exploratory analysis of semantic burden on 600 balanced-core records is reported in Appendix~\ref{app:v6-burden}.

\subsection{RQ1: Capability across Clinical Workflows}
\label{sec:rq1-capability}
\begin{table*}[t]
\caption{\textbf{Agent performance on \benchmark{} (\%).} A: \finalacc{}; S: \strictpass{}; E: \execsuccess{}. Overall is the macro-average across four patient-time scopes. Bold marks each group's highest displayed overall A and S, including ties.}
\label{tab:main-results}
\centering
\begingroup
\definecolor{resultnavy}{HTML}{264653}
\definecolor{resultteal}{HTML}{207D73}
\definecolor{resultgray}{HTML}{626C78}
\definecolor{resultband}{HTML}{EDF3F5}
\fontfamily{lmr}\fontsize{8}{9.1}\selectfont
\setlength{\tabcolsep}{1.5pt}
\setlength{\fboxsep}{2pt}
\renewcommand{\arraystretch}{1.0}
\newcommand{\resE}[1]{\textcolor{resultgray}{#1}}
\newcommand{\resA}[1]{\textcolor{resultnavy}{#1}}
\newcommand{\resS}[1]{\textcolor{resultteal}{#1}}
\newcommand{\NA}{\multicolumn{1}{c}{\textcolor{gray}{\textemdash}}}
\begin{tabular*}{\textwidth}{@{\extracolsep{\fill}}l*{15}{r}@{}}
\toprule
 & \multicolumn{3}{c}{\textbf{Overall}} & \multicolumn{3}{c}{Whole patient} & \multicolumn{3}{c}{Admission} & \multicolumn{3}{c}{ICU stay} & \multicolumn{3}{c@{}}{Event / study} \\
\cmidrule(lr){2-4}\cmidrule(lr){5-7}\cmidrule(lr){8-10}\cmidrule(lr){11-13}\cmidrule(l){14-16}
Agent / harness & \resA{A} & \resS{S} & \resE{E} & A & S & \resE{E} & A & S & \resE{E} & A & S & \resE{E} & A & S & \resE{E} \\
\midrule
\multicolumn{16}{@{}l@{}}{\rlap{\colorbox{resultband}{\parbox{\dimexpr\textwidth-2\fboxsep\relax}{\textcolor{resultnavy}{\textbf{GPT-5.6-Sol}}\hfill\textcolor{resultgray}{Coding harness}}}}} \\
Codex & \resA{\textbf{54.0}} & \resS{\textbf{29.4}} & \resE{86.4} & 40.6 & 18.8 & \resE{90.6} & 62.5 & 40.6 & \resE{100.0} & 61.3 & 19.4 & \resE{96.8} & 51.6 & 38.7 & \resE{58.1} \\
\multicolumn{16}{@{}l@{}}{\rlap{\colorbox{resultband}{\parbox{\dimexpr\textwidth-2\fboxsep\relax}{\textcolor{resultnavy}{\textbf{DeepSeek-V4-Pro}}\hfill\textcolor{resultgray}{Historical results$^{\dagger}$}}}}} \\
NoDataGuess & \resA{3.1} & \NA & \NA& 0.0 & \NA & \NA & 12.5 & \NA & \NA & 0.0 & \NA & \NA & 0.0 & \NA & \NA \\
ReAct & \resA{1.6} & \resS{0.0} & \resE{48.5} & 0.0 & 0.0 & \resE{43.8} & 3.1 & 0.0 & \resE{46.9} & 0.0 & 0.0 & \resE{41.9} & 3.2 & 0.0 & \resE{61.3} \\
DataVoyager & \resA{\textbf{6.3}} & \resS{\textbf{1.6}} & \resE{60.3} & 6.2 & 3.1 & \resE{81.2} & 9.4 & 0.0 & \resE{43.8} & 0.0 & 0.0 & \resE{45.2} & 9.7 & 3.2 & \resE{71.0} \\
Reflexion & \resA{\textbf{6.3}} & \resS{0.0} & \resE{85.0} & 6.2 & 0.0 & \resE{87.5} & 15.6 & 0.0 & \resE{78.1} & 0.0 & 0.0 & \resE{77.4} & 3.2 & 0.0 & \resE{96.8} \\
SelfDebug & \resA{1.6} & \resS{0.0} & \resE{45.3} & 3.1 & 0.0 & \resE{34.4} & 3.1 & 0.0 & \resE{46.9} & 0.0 & 0.0 & \resE{32.3} & 0.0 & 0.0 & \resE{67.7} \\
RAG-ReAct & \resA{3.1} & \resS{0.8} & \resE{44.6} & 6.2 & 3.1 & \resE{40.6} & 6.2 & 0.0 & \resE{31.2} & 0.0 & 0.0 & \resE{38.7} & 0.0 & 0.0 & \resE{67.7} \\
\multicolumn{16}{@{}l@{}}{\rlap{\colorbox{resultband}{\parbox{\dimexpr\textwidth-2\fboxsep\relax}{\textcolor{resultnavy}{\textbf{GLM-5.2}}\hfill\textcolor{resultgray}{Historical results$^{\dagger}$}}}}} \\
NoDataGuess & \resA{3.9} & \NA & \NA & 0.0 & \NA & \NA & 15.6 & \NA & \NA & 0.0 & \NA & \NA & 0.0 & \NA & \NA \\
ReAct & \resA{13.4} & \resS{0.8} & \resE{88.1} & 15.6 & 3.1 & \resE{84.4} & 21.9 & 0.0 & \resE{87.5} & 0.0 & 0.0 & \resE{90.3} & 16.1 & 0.0 & \resE{90.3} \\
DataVoyager & \resA{11.8} & \resS{2.4} & \resE{94.4} & 12.5 & 3.1 & \resE{100.0} & 18.8 & 3.1 & \resE{90.6} & 0.0 & 0.0 & \resE{93.5} & 16.1 & 3.2 & \resE{93.5} \\
Reflexion & \resA{\textbf{15.0}} & \resS{\textbf{3.2}} & \resE{98.4} & 12.5 & 3.1 & \resE{96.9} & 28.1 & 3.1 & \resE{100.0} & 0.0 & 0.0 & \resE{96.8} & 19.4 & 6.5 & \resE{100.0} \\
SelfDebug & \resA{13.4} & \resS{0.8} & \resE{92.8} & 12.5 & 3.1 & \resE{96.9} & 25.0 & 0.0 & \resE{96.9} & 0.0 & 0.0 & \resE{83.9} & 16.1 & 0.0 & \resE{93.5} \\
RAG-ReAct & \resA{10.2} & \resS{0.8} & \resE{88.1} & 12.5 & 3.1 & \resE{90.6} & 18.8 & 0.0 & \resE{87.5} & 0.0 & 0.0 & \resE{83.9} & 9.7 & 0.0 & \resE{90.3} \\
\multicolumn{16}{@{}l@{}}{\rlap{\colorbox{resultband}{\parbox{\dimexpr\textwidth-2\fboxsep\relax}{\textcolor{resultnavy}{\textbf{GPT-5.6-Luna}}\hfill\textcolor{resultgray}{Standard agents and coding harnesses}}}}} \\
NoDataGuess & \resA{3.9} & \NA & \NA & 0.0 & \NA & \NA & 15.6 & \NA & \NA & 0.0 & \NA & \NA & 0.0 & \NA & \NA \\
ReAct & \resA{13.4} & \resS{5.5} & \resE{93.7} & 12.5 & 6.2 & \resE{90.6} & 21.9 & 9.4 & \resE{87.5} & 0.0 & 0.0 & \resE{100.0} & 19.4 & 6.5 & \resE{96.8} \\
DataVoyager & \resA{11.9} & \resS{3.2} & \resE{72.3} & 9.4 & 3.1 & \resE{68.8} & 18.8 & 3.1 & \resE{65.6} & 0.0 & 0.0 & \resE{83.9} & 19.4 & 6.5 & \resE{71.0} \\
Reflexion & \resA{15.0} & \resS{\textbf{7.9}} & \resE{86.6} & 15.6 & 9.4 & \resE{84.4} & 25.0 & 9.4 & \resE{75.0} & 0.0 & 0.0 & \resE{93.5} & 19.4 & 12.9 & \resE{93.5} \\
SelfDebug & \resA{15.0} & \resS{\textbf{7.9}} & \resE{92.1} & 18.8 & 12.5 & \resE{84.4} & 21.9 & 3.1 & \resE{93.8} & 0.0 & 0.0 & \resE{96.8} & 19.4 & 16.1 & \resE{93.5} \\
RAG-ReAct & \resA{14.2} & \resS{\textbf{7.9}} & \resE{92.9} & 15.6 & 12.5 & \resE{90.6} & 25.0 & 6.2 & \resE{93.8} & 0.0 & 0.0 & \resE{93.5} & 16.1 & 12.9 & \resE{93.5} \\
Codex & \resA{15.0} & \resS{5.5} & \resE{100.0} & 18.8 & 15.6 & \resE{100.0} & 21.9 & 6.2 & \resE{100.0} & 0.0 & 0.0 & \resE{100.0} & 19.4 & 0.0 & \resE{100.0} \\
Claude Code & \resA{\textbf{21.3}} & \resS{\textbf{7.9}} & \resE{69.9} & 34.4 & 12.5 & \resE{56.2} & 28.1 & 3.1 & \resE{71.9} & 12.9 & 9.7 & \resE{80.6} & 9.7 & 6.5 & \resE{71.0} \\
\multicolumn{16}{@{}l@{}}{\rlap{\colorbox{resultband}{\parbox{\dimexpr\textwidth-2\fboxsep\relax}{\textcolor{resultnavy}{\textbf{Gemini-3.7-Flash}}\hfill\textcolor{resultgray}{Standard agents}}}}} \\
NoDataGuess & \resA{4.7} & \NA & \NA & 0.0 & \NA & \NA & 18.8 & \NA & \NA & 0.0 & \NA & \NA & 0.0 & \NA & \NA \\
ReAct & \resA{16.6} & \resS{4.7} & \resE{98.4} & 18.8 & 9.4 & \resE{96.9} & 28.1 & 3.1 & \resE{100.0} & 0.0 & 0.0 & \resE{100.0} & 19.4 & 6.5 & \resE{96.8} \\
DataVoyager & \resA{16.6} & \resS{\textbf{7.9}} & \resE{100.0} & 18.8 & 15.6 & \resE{100.0} & 28.1 & 9.4 & \resE{100.0} & 0.0 & 0.0 & \resE{100.0} & 19.4 & 6.5 & \resE{100.0} \\
Reflexion & \resA{\textbf{18.1}} & \resS{7.1} & \resE{99.2} & 18.8 & 12.5 & \resE{96.9} & 34.4 & 9.4 & \resE{100.0} & 0.0 & 0.0 & \resE{100.0} & 19.4 & 6.5 & \resE{100.0} \\
SelfDebug & \resA{16.6} & \resS{7.1} & \resE{99.2} & 15.6 & 12.5 & \resE{96.9} & 31.2 & 9.4 & \resE{100.0} & 0.0 & 0.0 & \resE{100.0} & 19.4 & 6.5 & \resE{100.0} \\
RAG-ReAct & \resA{15.8} & \resS{6.3} & \resE{100.0} & 15.6 & 15.6 & \resE{100.0} & 28.1 & 3.1 & \resE{100.0} & 0.0 & 0.0 & \resE{100.0} & 19.4 & 6.5 & \resE{100.0} \\
\multicolumn{16}{@{}l@{}}{\rlap{\colorbox{resultband}{\parbox{\dimexpr\textwidth-2\fboxsep\relax}{\textcolor{resultnavy}{\textbf{Qwen-3.8-Max}}\hfill\textcolor{resultgray}{Incomplete coverage$^{\ddagger}$}}}}} \\
NoDataGuess & \resA{0.0} & \NA & \NA & 0.0 & \NA & \NA & 0.0 & \NA & \NA & 0.0 & \NA & \NA & 0.0 & \NA & \NA \\
ReAct & \resA{\textbf{15.0}} & \resS{\textbf{8.0}} & \resE{61.3} & 12.5 & 6.2 & \resE{100.0} & 28.1 & 6.2 & \resE{100.0} & 0.0 & 0.0 & \resE{22.6} & 19.4 & 19.4 & \resE{22.6} \\
DataVoyager & \resA{13.4} & \resS{6.3} & \resE{50.3} & 18.8 & 15.6 & \resE{93.8} & 18.8 & 0.0 & \resE{71.9} & 0.0 & 0.0 & \resE{19.4} & 16.1 & 9.7 & \resE{16.1} \\
Reflexion & \resA{10.3} & \resS{4.8} & \resE{58.9} & 9.4 & 6.2 & \resE{96.9} & 18.8 & 0.0 & \resE{100.0} & 0.0 & 0.0 & \resE{19.4} & 12.9 & 12.9 & \resE{19.4} \\
SelfDebug & \resA{12.7} & \resS{6.4} & \resE{59.7} & 9.4 & 6.2 & \resE{100.0} & 21.9 & 3.1 & \resE{100.0} & 0.0 & 0.0 & \resE{19.4} & 19.4 & 16.1 & \resE{19.4} \\
RAG-ReAct & \resA{12.6} & \resS{7.2} & \resE{56.6} & 12.5 & 9.4 & \resE{93.8} & 21.9 & 3.1 & \resE{93.8} & 0.0 & 0.0 & \resE{19.4} & 16.1 & 16.1 & \resE{19.4} \\
\multicolumn{16}{@{}l@{}}{\rlap{\colorbox{resultband}{\parbox{\dimexpr\textwidth-2\fboxsep\relax}{\textcolor{resultnavy}{\textbf{Biomedical-agent transfers}}\hfill\textcolor{resultgray}{GPT-5.6-Luna backbone}}}}} \\
BioMedAgent & \resA{10.3} & \resS{3.1} & \resE{52.4} & 15.6 & 12.5 & \resE{59.4} & 12.5 & 0.0 & \resE{40.6} & 0.0 & 0.0 & \resE{45.2} & 12.9 & 0.0 & \resE{64.5} \\
BixBench & \resA{0.8} & \resS{0.8} & \resE{7.1} & 0.0 & 0.0 & \resE{9.4} & 3.1 & 3.1 & \resE{9.4} & 0.0 & 0.0 & \resE{3.2} & 0.0 & 0.0 & \resE{6.5} \\
BioMaster & \resA{0.0} & \resS{0.0} & \resE{0.8} & 0.0 & 0.0 & \resE{0.0} & 0.0 & 0.0 & \resE{0.0} & 0.0 & 0.0 & \resE{0.0} & 0.0 & 0.0 & \resE{3.2} \\
EHRAgent & \resA{\textbf{11.9}} & \resS{\textbf{6.4}} & \resE{87.4} & 15.6 & 12.5 & \resE{87.5} & 12.5 & 0.0 & \resE{78.1} & 0.0 & 0.0 & \resE{96.8} & 19.4 & 12.9 & \resE{87.1} \\
Biomni & \resA{0.0} & \resS{0.0} & \resE{0.0} & 0.0 & 0.0 & \resE{0.0} & 0.0 & 0.0 & \resE{0.0} & 0.0 & 0.0 & \resE{0.0} & 0.0 & 0.0 & \resE{0.0} \\
\bottomrule
\end{tabular*}
\par\vspace{3pt}
\begin{minipage}{\textwidth}\fontfamily{lmr}\fontsize{8}{9.1}\selectfont
$^{\dagger}$DeepSeek/GLM retain evaluator-compatible historical results. $^{\ddagger}$Qwen counts unevaluated tasks as failures. These groups are excluded from V6 diagnostics; unexecuted projections are omitted.
\end{minipage}
\endgroup
\end{table*}

GPT-5.6-Sol with Codex is the strongest observed combined system, reaching 54.0\% \finalacc{}, 29.4\% \strictpass{}, and 86.4\% \execsuccess{} (68 correct answers and 37 strict passes). This comparison does not isolate a harness effect. Complete Luna/Gemini standard configurations reach at most 7.9\% strict accuracy, and scaffold rankings vary by backbone and metric. Among biomedical adapters, EHRAgent leads with 11.9\% answer accuracy and 6.4\% strict accuracy; these are shared-interface transfer results, not rankings of unmodified native systems (Appendix~\ref{app:supplementary-diagnostics}).

\begin{figure}[t]
\centering
\includegraphics[width=\linewidth]{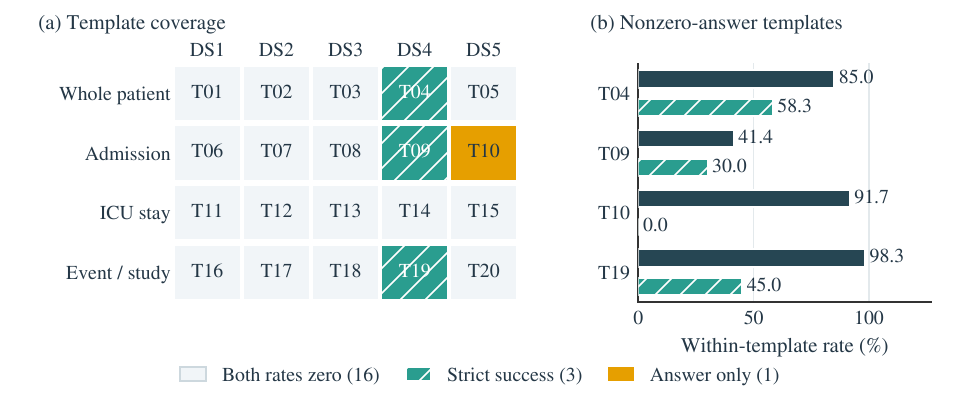}
\caption{\textbf{Performance across patient-time scopes and clinical capabilities.}
(a) Capability matrix and (b) template-level \finalacc{} and \strictpass{},
pooled across ten complete configurations.}
\label{fig:rq1-capability}
\end{figure}

Figure~\ref{fig:rq1-capability} summarizes template-level performance across the ten complete configurations (1,260 task--configuration records). In panel (a), DS1--DS5 follow the clinical capability taxonomy; gray cells indicate zero answer accuracy, hatching indicates nonzero strict success, and amber cells indicate answer-only success. Panel (b) plots percentage \finalacc{} and \strictpass{} in navy and teal.

Success is concentrated in four templates. T04, T09, T10, and T19 achieve answer/strict rates of 85.0/58.3\%, 41.4/30.0\%, 91.7/0.0\%, and 98.3/45.0\%; the other 16 templates have no correct answers. T14 has no correct predictions; T10 yields high answer accuracy without a valid workflow.

\subsection{RQ2: Where Clinical Workflows First Fail}
\label{sec:rq2-errors}
\begin{figure}[t]
\centering
\includegraphics[width=\linewidth]{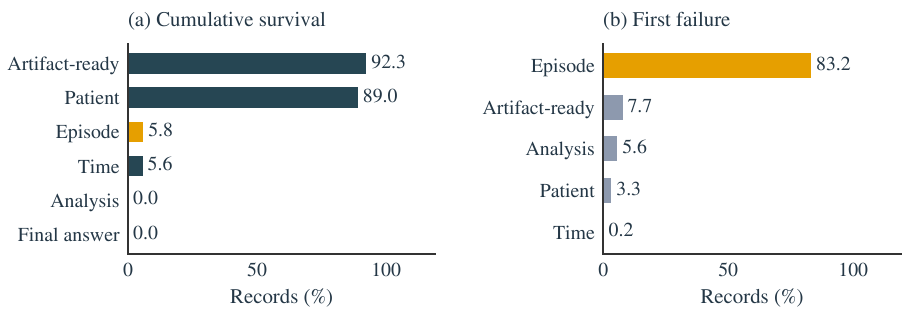}
\caption{\textbf{Workflow validation and first-failure distribution.}
(a) Cumulative gate survival and (b) mutually exclusive first failures
across 1,010 non-prediction task--configuration records.
Gates follow evaluator order.}
\label{fig:rq2-funnel}
\end{figure}

We localize failures through an ordered cumulative funnel: Artifact-ready $\rightarrow$ Patient $\rightarrow$ Episode $\rightarrow$ Time $\rightarrow$ Analysis $\rightarrow$ Final answer. It covers 1,010 non-prediction records because exported prediction reports omit the full grounding sequence. Figure~\ref{fig:rq2-funnel} shows cumulative gate survival in panel (a) and mutually exclusive first failures in panel (b). The first gate uses Artifact-ready, not \execsuccess{}. First failures follow evaluator order, not independent causes.

Episode grounding is the largest observed bottleneck. Of 1,010 records, 932 (92.3\%) are artifact-ready and 899 (89.0\%) pass patient grounding, but only 59 (5.8\%) also pass episode grounding. Episode selection accounts for 840 first failures (83.2\%), an 83.2-percentage-point drop in cumulative survival. No record passes the analysis stage, yielding zero cumulative final-answer success despite 5.4\% marginal answer accuracy. Inspected errors include wrong ECHO adjacency and duplicate CXR-to-admission links (Appendix~\ref{app:v6-diagnostics}). Such linkage errors change the analytical population before estimation. Few first failures at later gates imply neither correct temporal alignment nor valid analysis: earlier failures can mask additional downstream defects in the ordered first-failure summary.

\subsection{RQ3: Correct Answers versus Valid Workflows}
\label{sec:rq4-answer-validity}
\begin{figure}[t]
\centering
\includegraphics[width=\linewidth]{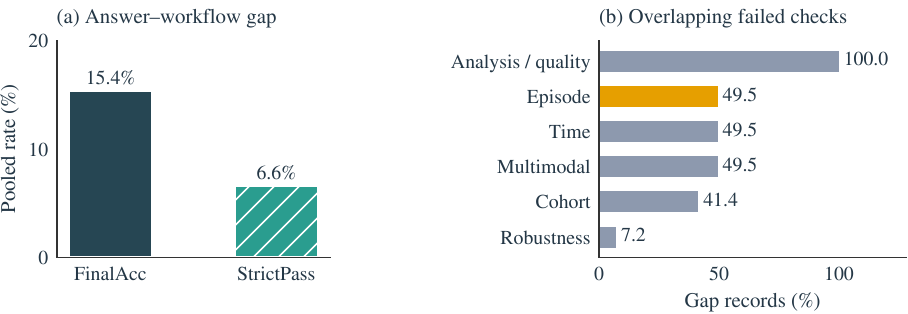}
\caption{\textbf{Final-answer correctness and workflow validity.}
(a) \finalacc{} versus \strictpass{} across 1,260 task--configuration records.
(b) Overlapping failed checks among 111 answer-correct but workflow-invalid submissions.}
\label{fig:rq4-gap}
\end{figure}

Only 83 of 194 correct answers (42.8\%) satisfy the full workflow contract. Across 1,260 records, pooled answer accuracy is 15.4\% and strict success is 6.6\%, leaving 111 answer-correct but invalid submissions (Figure~\ref{fig:rq4-gap}, panel (a)). Their mutually exclusive first failures divide into 55 episode-grounding failures, 51 prediction-quality failures, and 5 prediction-protocol failures.

Panel (b) reports overlapping failed checks among these 111 submissions: submissions can fail multiple checks, so percentages cannot be summed or read as independent defects. A target answer can match despite an incorrect event population or inadequate supporting model; the checks do not distinguish chance agreement from an answer insensitive to the upstream error (Appendix~\ref{app:gap-details}).

\begin{figure}[t]
  \centering
  \includegraphics[width=\linewidth]{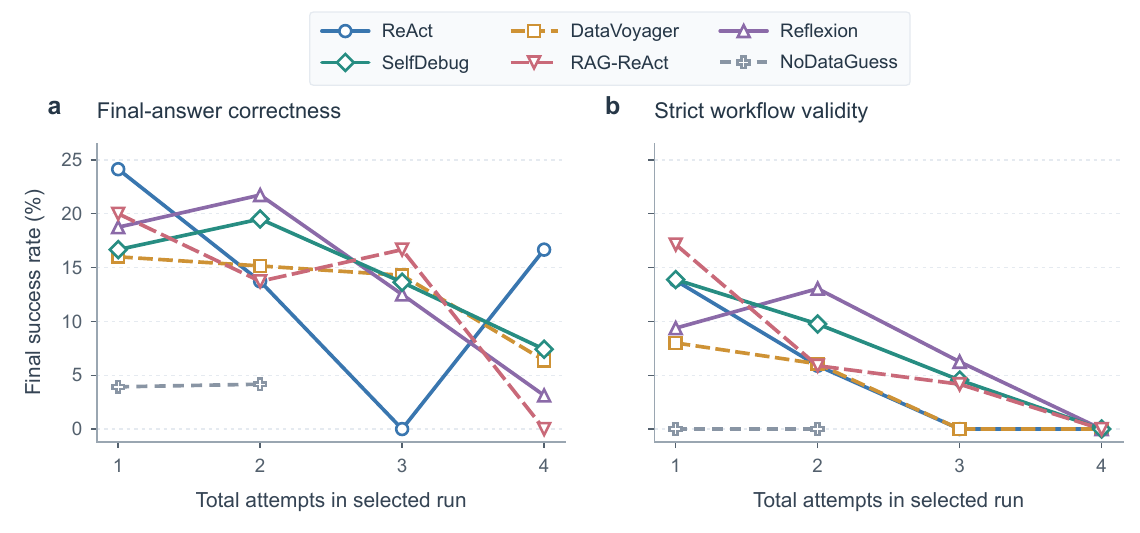}
\caption{\textbf{Performance by observed attempt count.}
GPT-5.6-Luna \finalacc{} (left) and \strictpass{} (right),
grouped by scaffold and total attempts.
Lines connect distinct task groups, not successive attempts on the same tasks.}
  \label{fig:rq5-attempt-performance}
\end{figure}

\subsection{RQ4: Attempt Counts and Performance--Efficiency Trade-offs}
\label{sec:rq5-efficiency}
We group GPT-5.6-Luna runs by their observed total attempts, using 126 tasks for each of six standard scaffolds (756 records). Each selected task--scaffold run enters one group. Figure~\ref{fig:rq5-attempt-performance} shows pooled final-answer correctness (left) and strict workflow validity (right) within each scaffold--attempt group. Lines connect distinct task groups, not successive scores for the same tasks; missing groups are left unplotted. Claude Code lacks comparable generation-and-repair attempt counts and is excluded.

More observed attempts do not consistently imply higher final success. Reflexion and SelfDebug have non-monotonic answer rates, and none of the five data-grounded scaffolds has a strict success in its four-attempt group. Task difficulty and stopping behavior can shape group composition. Without intermediate scores, the curves cannot establish whether revision improves or harms the same task.

Recorded computation likewise does not consistently accompany greater validity. SelfDebug reaches 14.99\% answer and 7.94\% strict accuracy at 46.02k tokens per task, close to ReAct's 45.68k. DataVoyager records 108.03k tokens with lower accuracy than ReAct. Claude Code reaches 21.27\% answer accuracy at 738.72k tokens but matches SelfDebug's strict rate. These are descriptive framework comparisons; missing auxiliary calls make some token counts lower bounds. Appendix~\ref{app:efficiency-details} provides the accounting protocol, per-group results, and the full framework efficiency table.

% \input{sections/discussion}
% \input{sections/limitations}
% !TEX root = ../main.tex

\section{Related Work}

\paragraph{General-purpose data science agents.}
General-purpose data science agents organize analytical work as an iterative process of planning, tool use, execution, and revision.
ReAct and Reflexion provide general mechanisms for action-feedback interaction and self-reflection, while DataVoyager applies agent-based reasoning to data-driven hypothesis generation and verification~\citep{yao2023react,
shinn2023reflexion,majumder2024datavoyager}.
Benchmarks examine complementary aspects of this process: DiscoveryBench evaluates data-driven discovery, ScienceAgentBench tests executable analyses derived from published studies, and BLADE assesses choices of variables, transformations, and statistical
models~\citep{majumder2024discoverybench,
chen2024scienceagentbench,gu2024blade}.
Evaluation also extends to multimodal scientific evidence in MoSciBench and longer research-replication workflows in PaperBench~\citep{liu2026moscibench,starace2025paperbench}.
These efforts establish important foundations for evaluating analytical execution.
\benchmark{} focuses on patient-centric clinical requirement: an executable analysis must use evidence belonging to the intended patient and episode, with asynchronously acquired modalities aligned to the task's temporal windows.

\paragraph{Medical data science agents.}
Medical data-science research adapts agentic analysis to
domain-specific records, tools, and scientific workflows,
with benchmarks assessing the resulting capabilities.
For clinical records, EHRAgent combines multi-table EHR
reasoning with iterative code execution~\citep{shi2024ehragent}.
For broader biomedical analysis, BioMedAgent and BioMaster
coordinate specialized agents and workflow stages, while
Biomni combines retrieval with programming and STELLA
develops reusable reasoning templates and expandable
tools~\citep{bu2026biomedagent,su2025biomaster,
huang2025biomni,jin2025stella}.
Evaluation accompanies these methodological advances:
BioDSBench and BixBench test biomedical programming and
computational analysis, and MedAgentGym supports both
evaluation and training through executable task
environments~\mbox{\citep{xu2026medagentgym}}.
Other clinical benchmarks emphasize modality breadth,
as in MultiMedBench, or explicit temporal evidence grounding,
as in CLIR-Bench~\citep{tu2023multimedbench,nie2026clir}.
Building on these complementary directions, \benchmark{}
jointly evaluates two requirements of longitudinal clinical
analysis: patient-centric grounding across admissions,
ICU stays, and repeated studies, and temporal alignment
of task-relevant EHRs, notes, ECGs, CXRs, and ECHOs.
Its artifact-level checks distinguish answer correctness
from the validity of the underlying workflow.
% Our transfer evaluation places adapted biomedical systems
% under this common interface, while
% Table~\ref{tab:related-work-clinical} summarizes selected
% differences in clinical modality coverage and evidence handling.
A comparison of clinical modalities and patient-centric
workflow coverage is provided in
Appendix~\ref{app:related-work},
Table~\ref{tab:related-work-clinical}.

% !TEX root = ../main.tex

\section{Conclusion}
We introduced \benchmark{}, a patient-centric benchmark for evaluating long-horizon LLM agents on longitudinal, multimodal clinical data science. Its 126 executable tasks make patient-episode grounding and asynchronous multimodal integration explicit requirements. Our experiments reveal substantial gaps between successful execution and valid clinical analysis. The strongest evaluated agent achieves 54.0\% final-answer accuracy but only a 29.4\% strict pass rate. Diagnostic analyses identify patient-episode grounding as the principal observed bottleneck. These findings highlight the construction and preservation of a valid patient-specific analytical context as a central challenge for clinical data science agents.

\clearpage
\bibliography{references}
\bibliographystyle{iclr2026_conference}

\appendix

\section{Task Construction and Operational Details}
\label{app:construction-details}
The following operational descriptions supplement the construction overview in the main text.

\paragraph{Task instructions.}
Each task is framed as a scientific question with three elements: (i) a patient-centered clinical context identifying the target patient and longitudinal setting; (ii) an analysis request specifying the clinical question, analysis population, temporal constraints, relevant modalities, and required computation; and (iii) an output contract defining the expected answer format and required artifacts. The clinical context and analysis request determine which patient episodes, time windows, and cross-modal evidence are relevant, while the output contract provides the anchor for evaluation. Final answers may be numerical values, categorical labels, short strings, or structured results, and the accompanying artifacts expose intermediate cohorts, temporal alignments, statistics, or model outputs needed to verify the analysis. Agents independently determine the necessary data exploration, preprocessing, cross-modal alignment, statistical analysis, and modeling steps.

\paragraph{Longitudinal multimodal patient packages.}
Each task provides a bounded, patient-first package that preserves both the longitudinal organization of a clinical trajectory and the native diversity of multimodal evidence. Records are organized by \texttt{subject\_id} and retain repeated hospital admissions (\texttt{hadm\_id}), ICU stays (\texttt{stay\_id}), and modality-specific events or studies (\texttt{study\_id}), together with their native timestamps. This hierarchy allows agents to reconstruct how a target patient's structured EHR events, clinical notes, ECGs, CXRs, and ECHOs evolve across episodes and time windows. The modalities also retain their original forms, including multi-table records, narrative text, physiological waveforms, radiographs, and echocardiographic studies. Modality availability, repeated examinations, and missing observations are preserved as properties of the source data. Schemas and file paths are standardized for access, while episode assignment for studies without encounter identifiers, cross-modal alignment, temporal filtering, repeated-measure aggregation, feature construction, and missing-data handling remain agent responsibilities rather than precomputed package properties. 

\paragraph{Evaluation.}
\benchmark{} adopts a patient-centered evaluation protocol: an agent is judged by whether its submission correctly reconstructs the target patient's longitudinal clinical context and produces an analytically valid answer. Each task is paired with an evaluator-private executable reference workflow that generates the expected artifacts and answer under fixed analytical and temporal specifications. Evaluation proceeds in three stages. First, the evaluator checks the existence, schema, and parseability of all required artifacts. Second, task-specific semantic checks validate target-patient identification, cohort construction, episode linkage, temporal boundaries, cross-modal alignment, and required statistical or model outputs. Third, the reported answer is compared with the reference result using declared tolerances for numerical outputs and task-specific matching rules for categorical or structured outputs. The deterministic references and task-specific matching rules make the operational evaluation automatic and reproducible across submissions to the same frozen task.

\paragraph{Patient-centered indexing and packaging.}
We integrate five PhysioNet resources: MIMIC-IV structured records~\citep{johnson2024mimiciv,johnson2023mimicivpaper}, MIMIC-IV-Note~\citep{johnson2023mimicnote}, MIMIC-IV-ECG~\citep{gow2023mimicecg}, MIMIC-CXR-JPG~\citep{johnson2024mimiccxrjpg}, and MIMIC-IV-ECHO~\citep{gow2026mimicecho}. Shared patient identifiers are used to assemble a bounded, multi-table package for each task, containing one named target patient and a reference cohort. Native patient, admission, ICU-stay, study, and event identifiers and timestamps are preserved, together with repeated examinations, one-to-many relationships, modality availability, and missing observations. Structured EHR events, clinical notes, ECG waveforms, chest radiographs, and echocardiographic studies retain their modality-specific representations and source clocks. We standardize schemas and file paths and register accessible raw assets, but do not precompute episode assignments, cross-modal joins, temporal alignments, features, or aggregate measurements. Agents use these packages to reconstruct the target patient's course and integrate multimodal evidence for the requested analysis.

\paragraph{Clinically grounded task design and instantiation.}
Task construction begins with realistic clinical data-analysis needs, such as characterizing a patient's longitudinal course, estimating associations or treatment effects, predicting subsequent outcomes from evidence available at a clinical landmark, and identifying recurring patterns in multimodal records. Each clinical need is translated into a patient-centered workflow that specifies the target-patient context, analysis population, eligibility criteria, temporal boundaries, relevant modalities, analytical objective, and expected outputs. We then instantiate six or seven eligible target patients per workflow template using fixed selection and sampling rules, independently materializing a patient package for each instance. An executable reference workflow operationalizes the clinical question under prespecified preprocessing, analytical, and fallback rules, confirming that the task is feasible and produces a reproducible result. The natural-language instruction, required artifacts, and evaluator are subsequently derived from this validated task protocol to maintain consistency between the intended clinical question and its evaluation.

\paragraph{Validation and quality control.}
Quality control is applied at both the template and instance levels. Clinical and statistical review examines the validity of cohort definitions, temporal directionality, modality use, variable definitions, and analytical procedures. Each materialized instance is then rerun in a clean environment, where automated checks verify package boundaries, modality and raw-asset accessibility, identifier and timestamp consistency, artifact schemas, row-level and temporal semantics, and agreement between the reference artifacts and final answer. Public task packages are separated from evaluator-private references to prevent access to gold artifacts or answers. Only instances whose packages, reference workflows, and expected submissions pass the complete evaluator are retained.

\paragraph{Illustrative workflows.}
The taxonomy requires agents to construct clinically and temporally valid analyses rather than retrieve static labels. For example, T09 constructs a 24-hour admission landmark and compares EHR-only and multimodal models for subsequent clinical deterioration. T18 uses a pretreatment CXR event to emulate a target trial comparing alternative pleural-drainage windows. T20 aligns CXR-derived pneumothorax evidence with respiratory support and procedures to discover reproducible acute management patterns. Section~\ref{sec:rq1-capability} reports results across all 20 templates; Appendix~\ref{app:v6-diagnostics} reports inspected failures showing how incorrect linkage changes the analysis units.

\paragraph{Reported allocation.}
The five analysis types in \benchmark{} capture complementary capabilities required for longitudinal clinical data science: (1) \emph{descriptive analysis}, which characterizes patient states and changes over time; (2) \emph{correlational analysis}, which quantifies relationships between clinical variables or modalities; (3) \emph{causal inference}, which estimates contrasts between clinically defined interventions or exposure strategies; (4) \emph{predictive modeling}, which predicts future outcomes using only evidence available before a specified landmark; and (5) \emph{pattern discovery}, which identifies reproducible trajectories, phenotypes, or clinical management patterns. Crossing these analysis types with the four patient-time levels produces 20 workflow templates; Figure~\ref{fig:taxonomy} illustrates one representative task at each patient-time level. The benchmark contains 26 descriptive tasks and 25 tasks in each of the other four categories, with six or seven independently materialized instances per template. This near-uniform allocation supports comparisons across both longitudinal scopes and analytical capabilities. Figure~\ref{fig:category-distribution} summarizes the five
clinical analysis capabilities and their task counts, complementing the breakdown of longitudinal scopes and represented modalities in Table~\ref{tab:coverage}.

\begin{table}[htbp]
\caption{\textbf{Summary of patient-time levels, represented modalities, and task counts in \benchmark{}.}
Modalities are aggregated across the workflow templates within each level; individual tasks use clinically relevant subsets.}
\label{tab:coverage}
\centering
\footnotesize
\renewcommand{\arraystretch}{1.1}
\begin{tabular}{@{}p{0.34\columnwidth}p{0.46\columnwidth}r@{}}
\toprule
\textbf{Patient-time level} &
\textbf{Represented modalities} &
\textbf{Tasks} \\
\midrule
Patient trajectory & EHR, NOTE, CXR, ECG, ECHO & 32 \\
Hospital admission & EHR, NOTE, CXR, ECG & 32 \\
ICU stay & EHR, NOTE, CXR, ECG & 31 \\
Clinical event/study & EHR, CXR, ECG, ECHO & 31 \\
\midrule
\multicolumn{2}{l}{\textbf{Total}} & \textbf{126} \\
\bottomrule
\end{tabular}
\end{table}

\section{Evaluation Protocol and Analysis Populations}
\label{app:evaluation-protocol}

\paragraph{Task suite and execution protocol.}
We evaluate the frozen V6 challenge suite of 126 tasks spanning 20 templates. The four patient-time levels contain 32, 32, 31, and 31 tasks. The suite emphasizes historically difficult workflows, so aggregate performance describes this stress suite rather than an IID distribution of clinical analyses. Standardized configurations use zero-shot prompting and temperature 0, with up to three generation or repair iterations and a maximum code-execution time of one hour per task; NoDataGuess receives one formatting iteration. Agents can access only their assigned package and manifest-listed raw assets. Private references, evaluator outputs, other packages, and previous submissions are withheld, restricting each run to its assigned task and permitted evidence.

\paragraph{Backbones and evidence coverage.}
The existing system comparison covers DeepSeek-V4-Pro, GLM-5.2, GPT-5.6-Luna, Gemini-3.7-Flash, and Qwen-3.8-Max. DeepSeek and GLM retain previously completed evaluator-compatible results because compatible sealed V6 Stage-B outputs are unavailable. Qwen uses the full 126-task denominator, counting tasks without evaluated submissions as unsuccessful. We therefore distinguish these descriptive comparisons from the complete V6 diagnostic cohort below. Unexecuted projected harness results are excluded from the empirical table.

\paragraph{Agent frameworks.}
We pair each backbone with six standardized agent settings. \textbf{NoDataGuess} receives the task instruction and answer schema without inspecting the patient package, providing an answer-prior control. \textbf{ReAct} alternates between reasoning, code execution, and observation~\citep{yao2023react}. \textbf{DataVoyager} first profiles the package and then invokes a planner--coder--critic workflow. \textbf{Reflexion} revises its analysis through self-reflection without reference feedback~\citep{shinn2023reflexion}. \textbf{SelfDebug} repairs generated programs using execution and submission-check traces~\citep{chen2023selfdebug}. \textbf{RAG-ReAct} retrieves task-relevant instructions and schema information from the assigned package before executing the reasoning, code-generation, and observation loop of a ReAct workflow.

We additionally evaluate Codex and Claude Code as interactive coding harnesses using GPT-5.6-Luna. These systems receive the same task instruction and patient package but use their native file-inspection, code-generation, execution, and revision loops. GPT-5.6-Sol with Codex is evaluated separately and reported as a model--harness combination, not an isolated harness comparison.

To examine the transfer of existing biomedical-agent designs, we adapt BioMedAgent~\citep{bu2026biomedagent}, the BixBench agent environment~\citep{mitchener2025bixbench}, BioMaster~\citep{su2025biomaster}, EHRAgent~\citep{shi2024ehragent}, and Biomni~\citep{huang2025biomni} to the same package and submission interface. The adapters preserve each system's planning, retrieval, programming, debugging, and checking components while connecting them to the \benchmark{} data and artifact contract. All five biomedical agents use GPT-5.6-Luna as their backbone under this shared interface.

\paragraph{Evaluation metrics.}
We report three complementary metrics. \finalacc{} (A) checks the designated final result using the task-specific categorical or numerical matching rule. \strictpass{} (S) additionally requires all applicable workflow gates:
\begin{equation}
\strictpass_i=\mathbb{1}\!\left[
C_i^{\mathrm{art}}\land C_i^{\mathrm{parse}}\land C_i^{\mathrm{rows}}
\land C_i^{\mathrm{val}}\land C_i^{\mathrm{answer}}\right],
\end{equation}
where the gates check artifact completeness, schema, cohort and row semantics, validation invariants, and final-answer correctness. The 24 robustness tasks must pass both base and additional analyses. For measured V6 prediction workflows, both EHR-only and multimodal models must satisfy
\begin{equation}
\mathrm{AUROC}\geq\max(0.50,\mathrm{AUROC}_{\mathrm{ref}}-0.10)
\end{equation}
and achieve AUPRC at least as high as hidden-test outcome prevalence. \execsuccess{} (E) records completed execution with a runner-accepted submission. Separately, \emph{Artifact-ready} checks completeness and schema from evaluator reports and differs from \execsuccess{}.

\paragraph{Analysis populations and aggregation.}
The new analyses reuse frozen task configurations, submissions, and evaluator reports without rerunning models or reference workflows. They cover ten complete data-grounded configurations: GPT-5.6-Luna and Gemini-3.7-Flash, each with ReAct, DataVoyager, Reflexion, SelfDebug, and RAG-ReAct. Qwen's incomplete runs, historical results, NoDataGuess, coding harnesses, and biomedical transfers are excluded from this diagnostic cohort. Table~\ref{tab:analysis-populations} specifies the analysis denominators. The main system and efficiency tables report unweighted macro-averages across four patient-time levels; diagnostic percentages pool the stated task--configuration records and should not be compared as identical aggregations.

\begin{table}[htbp]
\caption{Analysis populations for the diagnostic evaluations. All four analyses use the same ten complete configurations; A1, A2, A3, and A6 are source-report identifiers.}
\label{tab:analysis-populations}
\centering\small
\begin{tabular}{llll}
\toprule
Analysis & Location & Included tasks & Records \\
\midrule
A1: capability matrix & RQ1 & All 126 & 1,260 \\
A2: semantic funnel & RQ2 & 101 non-prediction tasks & 1,010 \\
A3: semantic burden & Appendix~\ref{app:v6-burden} & I01--I03 of each template & 600 \\
A6: answer--workflow gap & RQ3 & All 126 & 1,260 \\
\bottomrule
\end{tabular}
\end{table}

\section{Supplementary Diagnostic Results}
\label{app:supplementary-diagnostics}

\subsection{System comparisons and transfer}
\paragraph{End-to-end performance remains limited.}
Table~\ref{tab:main-results} reports system-level results. GPT-5.6-Sol with Codex is the strongest observed system, with 54.0\% \finalacc{}, 29.4\% \strictpass{}, and 86.4\% \execsuccess{} under patient-time macro averaging. This corresponds to 68 answer-correct tasks and 37 strict passes out of 126. These results describe the combined model and harness, rather than an isolated harness effect. In the standardized grid, Gemini-3.7-Flash with Reflexion attains the highest \finalacc{} (18.1\%). The highest reported \strictpass{} is 8.0\% for Qwen with ReAct, subject to the coverage caveat above; the complete Luna and Gemini configurations reach 7.9\%.

\paragraph{Data access helps, but scaffold rankings depend on the metric.}
NoDataGuess obtains 0.0--4.7\% \finalacc{} and no strict passes. Within the complete V6 cohort, data-grounded scaffolds improve answer accuracy, but none consistently dominates across backbones and metrics. Gemini with Reflexion leads on \finalacc{}, whereas DataVoyager leads on \strictpass{} for that backbone. Under Luna, Claude Code attains 21.3\% \finalacc{} and 7.9\% \strictpass{}, compared with Codex's 15.0\% and 5.5\%, even though execution success is lower (69.9\% versus 100\%). Thus, successful execution alone does not establish a valid clinical workflow or preserve strict-validity rankings.

\paragraph{The capability breakdown localizes success to four templates.}
Figure~\ref{fig:rq1-capability} crosses patient-time levels with five analysis capabilities for the ten complete configurations. Only T04, T09, T10, and T19 have nonzero \finalacc{}, at 85.0\%, 41.4\%, 91.7\%, and 98.3\%, respectively; their strict rates are 58.3\%, 30.0\%, 0.0\%, and 45.0\%. All remaining 16 templates have zero final-answer accuracy. Prediction is not uniformly tractable: T14, the ICU-stay prediction template, has no successful answer. T10 exhibits the opposite mismatch, with high pattern-label accuracy but no validated workflow. Capability and patient-time level jointly distinguish failures hidden by aggregate accuracy.

\paragraph{Biomedical-agent transfer remains selective.}
Among five adapted systems using Luna, EHRAgent performs best, reaching 11.9\% \finalacc{}, 6.4\% \strictpass{}, and 87.4\% \execsuccess{}. BioMedAgent obtains 10.3\% and 3.1\% under the two correctness metrics, BixBench obtains 0.8\% under each, and BioMaster and Biomni obtain no correct answers. None answers an ICU-stay task correctly. These results describe interface transfer by the evaluated adapters, not rankings of the original systems under their native data, tools, and evaluation protocols.

\subsection{Failure localization}
\label{app:v6-diagnostics}
\paragraph{An ordered funnel separates usable artifacts from valid analysis units.}
We construct the cumulative sequence Artifact-ready $\rightarrow$ Patient $\rightarrow$ Episode $\rightarrow$ Time $\rightarrow$ Analysis $\rightarrow$ Final answer. A record survives a stage only if it passes that stage and all preceding ones. Analysis aggregates downstream cohort, raw-evidence, and analytical requirements. We restrict the funnel to 1,010 non-prediction records because exported prediction Stage-B reports do not expose the full patient/episode/time gate sequence. First failures follow evaluator order and do not independently count underlying errors.

\paragraph{The largest drop occurs at episode grounding.}
Of 1,010 records, 932 (92.3\%) are artifact-ready and 899 (89.0\%) pass patient grounding, but only 59 (5.8\%) also pass episode grounding (Figure~\ref{fig:rq2-funnel}). Episode selection is the first failure in 840 records (83.2\%), compared with 78 artifact failures, 57 analysis failures, 33 patient failures, and two time-window failures. Only 57 records reach the analysis stage, and none passes it. The 83.2-percentage-point drop from patient to episode survival identifies the principal observed bottleneck as reconstructing admissions, ICU stays, studies, and eligible event units. The small number of first time-window failures does not imply that temporal alignment is easy: most records have already failed an earlier gate and cannot first fail here.

\paragraph{Template-level evidence identifies concrete reconstruction errors.}
Among 950 records from the 15 zero-accuracy non-prediction templates, 785 (82.6\%) first fail episode grounding. The report's inspected examples include choosing non-adjacent ECHO studies instead of applying the specified timestamp and study-ID tie-break (T02), linking the same CXR study to multiple admissions (T03), and duplicating episode identities in 33/60 T06 submissions. These errors change the analytical population or exposure window before a statistical estimate is computed. T14 requires separate interpretation: 59/60 prediction records first fail exact private-test-unit coverage and one fails Stage-B execution. These are inspected failures, not a complete per-template diagnostic report.

\subsection{Semantic Burden and Workflow Success}
\label{app:v6-burden}
\label{sec:rq3-burden}

\paragraph{Motivation and analysis population.}
We examine whether semantic grounding burden is associated with workflow success after adjustment for observed task characteristics. This exploratory analysis uses I01--I03 from each of the 20 templates, yielding 60 tasks and 600 task--configuration records. The ten complete configurations pair GPT-5.6-Luna and Gemini-3.7-Flash with ReAct, DataVoyager, Reflexion, SelfDebug, and RAG-ReAct.

\begin{figure}[t]
\centering
\includegraphics[width=\linewidth]{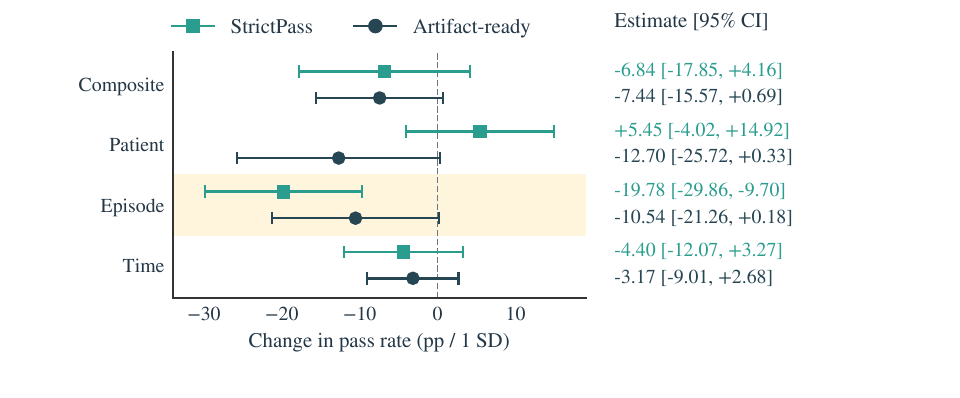}
\caption{\textbf{Episode burden is negatively associated with strict success.} Adjusted coefficients with 95\% confidence intervals for \strictpass{} and Artifact-ready.}
\label{fig:rq3-burden}
\end{figure}

\paragraph{Burden annotations and estimation.}
Contract-based annotations cover patient burden (repeated units, reference fitting, and patient-level splits or clustering), episode burden (admission/stay/study joins and multiple episodes), and time burden (windows, ordering, landmarks, future outcomes, tie-breaks, availability constraints, and resampling). We estimate linear probability models controlling for agent configuration, patient-time level, capability, package bytes, and file count, with standard errors clustered by the 20 templates. Exact CSV row counts were unavailable and are not included as task-level controls. Figure~\ref{fig:rq3-burden} reports adjusted coefficients and 95\% confidence intervals in percentage points per one-standard-deviation increase in burden. Squares denote \strictpass{} and circles denote Artifact-ready, which checks artifact completeness and schema rather than execution success.

\paragraph{Episode burden is negatively associated with strict success.}
The episode-burden coefficient for \strictpass{} is $-19.78$ percentage points per standard deviation (95\% CI $[-29.86,-9.70]$, $p=0.0001$). Patient and time estimates are $+5.45$ and $-4.40$ points, respectively, with both confidence intervals including zero. The episode-specific association is consistent with the failure-localization results in Section~\ref{sec:rq2-errors}.

\paragraph{Composite burden does not establish an independent association.}
The composite-burden coefficient for \strictpass{} is $-6.84$ percentage points (95\% CI $[-17.85,4.16]$, $p=0.2229$). Its coefficient for Artifact-ready is $-7.44$ points (95\% CI $[-15.57,0.69]$, $p=0.0729$). Both intervals include zero, so these estimates do not establish an independent association between composite burden and either outcome after the stated adjustment.

\paragraph{Interpretation and limitations.}
These associations are exploratory rather than causal: burden varies across templates, not through a randomized intervention. Unobserved template difficulty, only 20 clusters, and annotations without an independent second rater limit interpretation. Matched task variants that alter episode-selection demands while preserving the analytical objective would provide a more controlled test of the episode-burden hypothesis.

\subsection{Answer--workflow gap}
\label{app:gap-details}
\paragraph{Two distinct mechanisms explain the gap.}
The earliest failures include 55 episode-grounding failures and 56 analysis/model-quality failures: 51 fail \texttt{PredictionQualityFloorPass} and five fail \texttt{PredictionProtocolPass}: the designated target prediction can match within tolerance even when model quality or the prediction protocol is inadequate. The 55 episode failures match the answer-correct count for T10, whose strict rate is zero. Agreement on a final label can therefore coexist with an incorrect event population; the checks alone cannot distinguish chance agreement from a label insensitive to upstream errors under the specified matching rule.

\paragraph{Overlapping checks expose coupled defects.}
All 111 gap records fail an analysis/model-quality check; 55 also fail each of episode grounding, time alignment, and multimodal evidence, 46 fail cohort construction, and eight fail robustness checks. These categories overlap and differ from the mutually exclusive first-failure partition. They show why summing failed gates exaggerates the number of independent mistakes: one incorrect event set can invalidate several subsequent checks.

\section{Attempt Counts and Recorded Resource Use}
\label{app:efficiency-details}

\paragraph{Comparison protocol.}
We compare the six standard scaffolds and Claude Code under GPT-5.6-Luna on the same 126 tasks. For each final submission, we aggregate the recorded input and output tokens of the selected run, including its recorded repair calls, and pair them with patient-time macro \finalacc{}, \strictpass{}, and elapsed time. Table~\ref{tab:rq5-framework-efficiency} reports these framework-level performance and efficiency values. Because some auxiliary calls are absent from the saved logs, the token values for DataVoyager, Reflexion, and Claude Code are recorded lower bounds; they support descriptive comparisons of recorded token use, not estimates of complete billing cost or total computation.

\paragraph{Attempt-count analysis.}
To examine how repeated attempts relate to final outcomes, we group the six standard scaffolds by the total recorded attempts in each selected run (Figure~\ref{fig:rq5-attempt-performance}). The analysis includes 756 task--framework records, with 126 tasks per scaffold. Within each framework, each task contributes once to the group matching its observed attempt count, from one to four. Each plotted rate is the fraction of tasks in that group whose final submission passes the corresponding metric; these pooled group rates differ from the patient-time macro rates in Table~\ref{tab:rq5-framework-efficiency}. NoDataGuess has observations only at one and two attempts. Claude Code is excluded because its reported interaction turns do not provide an attempt count comparable to the generation and repair attempts of standard scaffolds.

\begin{table}[htbp]
\caption{GPT-5.6-Luna framework efficiency on the same 126 tasks. \finalacc{} and \strictpass{} are patient-time macro rates; token and time values are task means. $^{*}$ marks incomplete token accounting.}
\label{tab:rq5-framework-efficiency}
\centering\small
\setlength{\tabcolsep}{5pt}
\begin{tabular}{lrrrr}
\toprule
Framework & \finalacc{} (\%) & \strictpass{} (\%) & Tokens (k) & Time (min) \\
\midrule
NoDataGuess & 3.91 & 0.00 & 16.09 & 1.02 \\
ReAct & 13.43 & 5.52 & 45.68 & 3.52 \\
DataVoyager & 11.87 & 3.18 & 108.03$^{*}$ & 5.65 \\
Reflexion & 14.99 & 7.91 & 48.47$^{*}$ & 4.53 \\
SelfDebug & 14.99 & 7.94 & 46.02 & 3.54 \\
RAG-ReAct & 14.19 & 7.91 & 45.17 & 6.42 \\
Claude Code & 21.27 & 7.94 & 738.72$^{*}$ & 16.35 \\
\bottomrule
\end{tabular}
\end{table}

\paragraph{More attempts do not consistently correspond to higher final success.}
Final-answer correctness varies non-monotonically across attempt groups for several frameworks. For example, Reflexion reaches 18.75\% in the one-attempt group and 21.74\% in the two-attempt group, followed by 12.50\% and 3.13\% in the three- and four-attempt groups. SelfDebug shows a similar initial difference, from 16.67\% to 19.51\%, followed by 13.64\% and 7.41\%. Across all five data-grounded scaffolds, the four-attempt groups contain no strict successes (0/18 for ReAct, 0/47 for DataVoyager, 0/32 for Reflexion, 0/27 for SelfDebug, and 0/16 for RAG-ReAct). Thus, runs with more observed attempts do not consistently end with higher answer correctness or workflow validity.

\paragraph{What the attempt-count comparison establishes.}
The curves describe final outcomes conditional on observed attempt counts. Task difficulty and framework stopping behavior can affect which tasks enter each group, and the saved data contain no intermediate scores. Consequently, these results cannot establish whether an additional revision improves or harms the same task. Measuring revision gains requires scoring intermediate submissions on a fixed task cohort with controlled revision budgets and stopping rules, rather than comparing groups with different task compositions.

\paragraph{Additional orchestration is not consistently efficient.}
DataVoyager records 108.03k tokens per task, 2.37$\times$ ReAct's 45.68k, but obtains lower final-answer accuracy (11.87\% versus 13.43\%) and lower strict accuracy (3.18\% versus 5.52\%). SelfDebug records 46.02k tokens per task, nearly identical to ReAct, while improving final-answer accuracy to 14.99\% and strict accuracy to 7.94\%. In micro efficiency terms, SelfDebug produces 3.28 answer-correct tasks and 1.72 strict passes per million recorded tokens, compared with 2.95 and 1.22 for ReAct and 1.10 and 0.29 for DataVoyager. Reflexion and RAG-ReAct reach similar accuracy to SelfDebug, but require more elapsed time: 4.53 and 6.42 minutes per task, respectively, compared with 3.54 minutes for SelfDebug.

\paragraph{The coding harness improves answers at substantially higher cost.}
Claude Code reaches the highest GPT-5.6-Luna final-answer accuracy, 21.27\%, but records 738.72k tokens and 16.35 minutes per task. This is 16.17$\times$ ReAct's recorded token use and 4.64$\times$ its elapsed time. Its \strictpass{} accuracy is 7.94\%, equal to SelfDebug at the reported precision despite the large resource difference. In this comparison, the harness obtains more answer-correct outputs at higher recorded cost, without a corresponding increase in fully validated workflows. Framework differences and incomplete token accounting prevent attributing the observed performance differences to additional computation alone.

\paragraph{Most recorded computation does not end in a successful submission.}
Across the twelve complete standard configurations of GPT-5.6-Luna and Gemini-3.7-Flash, the selected runs contain 91.79 million recorded tokens. Final-answer failures account for 85.80\% of these tokens, and strict failures account for 94.46\%. This distribution mirrors RQ2: agents can spend substantial computation producing and repairing artifacts while retaining an incorrect episode, cohort, or analytical result.

\section{Evaluation Design and Agent-Design Implications}
\label{app:discussion-details}

\paragraph{What the benchmark currently measures.}
The present release stresses multimodal data engineering and analysis more than end-to-end perceptual model training. Packages expose raw CXR, ECG, and ECHO assets for audit and optional analysis, but also include structured labels, measurements, and metadata so that a task is feasible under bounded compute. The observed failures therefore cannot be attributed mainly to image or waveform recognition. They instead reveal difficulty with linking modality events to encounters, enforcing cutoffs, constructing cohorts, selecting statistical methods, and extracting the requested answer. Future benchmark tracks can isolate raw-perception dependence by withholding derived labels and measurements but preserving raw-asset access.

\paragraph{Why deterministic answers remain useful.}
Open clinical analyses can admit multiple defensible workflows, but unrestricted judging is expensive and difficult to reproduce. We retain open implementation choices while fixing the operational definitions that determine the scored quantity. Intermediate artifacts make disagreements diagnosable. This design is intentionally stricter than a narrative answer judge: a plausible clinical interpretation cannot compensate for a leaked target window or a wrong AUROC difference. A calibrated hierarchical rubric could later award credit for valid intermediate steps while retaining the deterministic endpoint for complete workflows.

\paragraph{Implications for agent design.}
The results support three priorities. First, agents need explicit representations of patient, admission, ICU, and event-time relations rather than treating all CSV files as interchangeable tables. Second, validation should recompute invariants, cohort counts, and temporal boundaries, not only inspect whether code exits. Third, agents should trace the final scalar to a specific artifact and row. The high cost and lower performance of DataVoyager suggest that additional planning layers should be conditional on detected uncertainty rather than always invoked.

\section{Limitations}
\label{app:limitations}

The benchmark is derived from a single health system and inherits the demographic, clinical-practice, documentation, and missingness biases of MIMIC-IV. It evaluates retrospective data analysis, not bedside diagnosis or treatment. The 126-task challenge suite deliberately oversamples difficult templates and is not an IID test set; results should be compared only on the same frozen manifest. Most experiments use one rollout per task, so Pass$^3$, variance estimates, and statistical significance remain to be measured. We have completed automated reference validation and manual realism review, but have not yet reported a blinded multi-clinician adjudication study or human data-scientist baseline. The medical-agent adapters necessarily modify I/O and sandboxes, and therefore test transfer rather than exact native-system performance. Proprietary snapshots and an API aggregation layer also limit rerun stability; releases must include request parameters, logs, and raw outputs.

% !TEX root = ../main.tex

\section{Comparison with Related Benchmarks}
\label{app:related-work}
Table~\ref{tab:related-work-clinical} compares the clinical
inputs and workflow requirements emphasized by related
benchmarks. We distinguish modality coverage from evidence
handling: providing multiple data sources does not necessarily
require agents to link observations to the correct patient,
episode, and time window. In \benchmark{}, agents work with
structured EHRs, clinical notes, CXRs, ECGs, and echocardiograms
within bounded patient-centered packages. The comparison
therefore considers not only which modalities are available,
but also whether temporal alignment and the handling of
missing or conflicting evidence are explicit task requirements.
Coverage labels describe benchmark emphasis rather than
exhaustive system capabilities.
\begin{table}[t]
  \centering
  \caption{\textbf{Clinical modality and workflow emphasis
  in related work.}
  The comparison focuses on clinical inputs and
  patient-centric evidence handling.}
  \label{tab:related-work-clinical}

  \begingroup
  \small
  \definecolor{rwgreen}{HTML}{35815C}
  \definecolor{rwred}{HTML}{BA403B}
  \definecolor{rwamber}{HTML}{AE730C}
  \definecolor{rwband}{HTML}{EEE9F5}

  \newcommand{\rwfull}{\textcolor{rwgreen}{\ding{51}}}
  \newcommand{\rwpartial}{\textcolor{rwamber}{\textbf{P}}}
  \newcommand{\rwnonprimary}{\textcolor{rwred}{\ding{55}}}

  \setlength{\tabcolsep}{2.5pt}
  \renewcommand{\arraystretch}{1.15}

  \begin{tabularx}{\linewidth}{@{}l
    *{5}{>{\hsize=.75\hsize
           \linewidth=\hsize
           \centering\arraybackslash}X}
    *{2}{>{\hsize=1.625\hsize
           \linewidth=\hsize
           \centering\arraybackslash}X}@{}}
    \toprule
    Work
      & EHR & Notes & CXR & ECG & ECHO
      & \shortstack{Temporal\\alignment}
      & \shortstack{Missingness/\\conflicts} \\
    \midrule

    % Verify work names and add the corresponding \citep{...}.
    CliniDSBench / BioDSBench
      & \rwpartial
      & \rwnonprimary
      & \rwnonprimary
      & \rwnonprimary
      & \rwnonprimary
      & \rwpartial
      & \rwpartial \\

    STELLA
      & \rwpartial
      & \rwnonprimary
      & \rwnonprimary
      & \rwnonprimary
      & \rwnonprimary
      & \rwpartial
      & \rwpartial \\

    MedAgentGym
      & \rwfull
      & \rwfull
      & \rwnonprimary
      & \rwnonprimary
      & \rwnonprimary
      & \rwpartial
      & \rwpartial \\

    \rowcolor{rwband}
    \textbf{\benchmark{} (ours)}
      & \rwfull
      & \rwfull
      & \rwfull
      & \rwfull
      & \rwfull
      & \rwfull
      & \rwfull \\
    \bottomrule
  \end{tabularx}

  \par\vspace{3pt}
  \begin{minipage}{\linewidth}
    \footnotesize\raggedright
    \rwfull{} primary or explicit coverage;
    \rwpartial{} partial, limited, or indirect coverage;
    \rwnonprimary{} non-primary emphasis, not demonstrated absence.
    Temporal alignment refers to cross-modal patient-time alignment.
    \textit{Draft comparison; work identities, citations,
    and coverage labels require source verification.}
  \end{minipage}
  \endgroup
\end{table}
% !TEX root = ../../main.tex

\section{Template Catalog}
\label{app:templates}
The 20 templates span four patient-time scopes and five analytical categories. Each template defines a reusable analysis specification, while individual task instances bind it to a target patient and a bounded data package. Table~\ref{tab:templates} lists representative scored outputs; full workflow validity additionally requires satisfying the artifact, temporal, and semantic requirements specified in the corresponding task cards.
\begin{table}[htbp]
\caption{The 20 benchmark templates and their scored outputs. Full task cards additionally specify temporal windows, package scope, required artifacts, validation rules, and tolerance.}
\label{tab:templates}
\centering
\scriptsize
\setlength{\tabcolsep}{2.8pt}
\begin{tabular}{clll}
\toprule
ID & Time scale & Analysis & Representative scored output \\
\midrule
T01 & Patient & Profiling & trajectory enrichment \\
T02 & Patient & Association & adjusted lag coefficient \\
T03 & Patient & Temporal effect & difference-in-differences effect \\
T04 & Patient & Prediction & multimodal AUROC gain \\
T05 & Patient & Phenotyping & transition enrichment \\
T06 & Admission & Profiling & onset-cascade separation \\
T07 & Admission & Association & adjusted association \\
T08 & Admission & Temporal effect & event-study effect \\
T09 & Admission & Prediction & landmark AUROC gain \\
T10 & Admission & Phenotyping & cluster-outcome enrichment \\
T11 & ICU stay & Profiling & early-burden enrichment \\
T12 & ICU stay & Association & distributed-lag odds ratio \\
T13 & ICU stay & Temporal effect & intervention response effect \\
T14 & ICU stay & Prediction & best modality combination \\
T15 & ICU stay & Phenotyping & phenotype enrichment \\
T16 & Event/study & Profiling & evidence-age degradation \\
T17 & Event/study & Association & CXR-anchored adjusted odds ratio \\
T18 & Event/study & Temporal effect & pre/post event effect \\
T19 & Event/study & Prediction & ECG-anchored AUROC gain \\
T20 & Event/study & Pattern mining & CXR-note discordance pattern \\
\bottomrule
\end{tabular}
\end{table}

% \input{sections/appendix/challenge_suite_composition}
% \input{sections/appendix/medical_agent_diagnostic}
% \input{sections/appendix/evaluator_checks}
% \input{sections/appendix/reproducibility_checklist}
% !TEX root = ../../main.tex

\section{Ethics, Data Governance, and Release}

All source datasets are deidentified PhysioNet resources~\citep{goldberger2000physionet} governed by credentialed-access agreements. The benchmark does not attempt reidentification and must not be used for direct patient care. Raw MIMIC records cannot be redistributed through an unrestricted repository. We will release construction code, task schemas, evaluator code, and non-sensitive metadata publicly, while distributing derived patient packages and selected raw assets through PhysioNet under the applicable credentialed health-data license. Development-only absolute paths and full-repository locators will be removed from release manifests. The evaluator blocks network access, manifest-external raw assets, reference and gold files, and scans outside the task package. Agent outputs that resemble diagnosis or treatment advice are flagged, because the intended task is retrospective computation over research records, not a clinical recommendation for patient care.

\end{document}